\documentclass[letterpaper, 10 pt, conference]{ieeeconf}  

\IEEEoverridecommandlockouts

\usepackage{amsmath} 
\usepackage{amssymb}  
\usepackage[dvipsnames]{xcolor}
\usepackage{xspace}
\usepackage{url}

\usepackage[linesnumbered,ruled,vlined]{algorithm2e}

\usepackage{graphicx} 

\usepackage{array}
\usepackage{booktabs}
\usepackage{adjustbox}

\usepackage{cite}

\usepackage{hyperref}
\usepackage{subfigure}
\usepackage{bytefield}
\usepackage{multirow}
\usepackage{tikz}
\usepackage{rotating}
\usetikzlibrary{positioning, shapes, arrows, automata}
\usepackage{glossaries}
\glsdisablehyper 
\usepackage[normalem]{ulem}
\usepackage{fancyhdr}
\usepackage{bbm}

\usepackage{marginnote}

\usepackage{fancyvrb} 

\usepackage[most]{tcolorbox}

\usepackage[frozencache=true,cachedir=minted-cache]{minted}

\usepackage{xcolor}
\usepackage{minted}

\usepackage{algpseudocode}

\fancypagestyle{withfooter}{
  
  \fancyfoot[C]{\footnotesize }
}

\hypersetup{
    colorlinks,
    linkcolor={red!50!black},
    citecolor={blue!50!black},
    urlcolor={blue!80!black}
}

\lstdefinelanguage{json}{
  basicstyle=\ttfamily,
  keywordstyle=\color{blue},
  stringstyle=\color{red},
  morestring=[b]",
  morecomment=[l]{//},
  morecomment=[s]{/*}{*/},
  morekeywords={true,false,null}
}

\newif\ifdraftcolor

\ifdraftcolor

\newcommand{\vm}[1]{{\color{teal}#1}}
\newcommand{\todo}[1]{{\color{red}#1}}

\newcommand{\deletecamready}[1]{{\color{Red}\sout{#1}}}

\newcommand{\tmpcomment}[1]{#1}

\newcommand{\changed}[1]{{\color{blue} #1}}
\else
\newcommand{\todo}[1]{}

\newcommand{\deletecamready}[1]{}

\newcommand{\vm}[1]{#1}

\newcommand{\tmpcomment}[1]{}
\newcommand{\changed}[1]{{#1}}
\fi

\newcommand{\eg}{\textit{e}.\textit{g}.,}

\newcommand{\citationneeded}[1]{$^{[\text{\color{blue}citation needed}]}$~}

\newcommand{\taskname}{incomplete}

\newcommand{\figvspace}{\vspace{-2mm}}
\newcommand{\tabvspace}{\vspace{-1mm}}
\newcommand{\shortskip}{\vspace{3pt}}

\title{{\LARGE \bf
SPINE: Online Semantic Planning for Missions with Incomplete Natural Language Specifications in Unstructured Environments} 
}

\author{Zachary Ravichandran, Varun Murali, Mariliza Tzes, George J. Pappas, and Vijay Kumar\thanks{All authors are with are with the GRASP Laboratory, University of Pennsylvania. Corresponding author: {\tt\small zacravi@seas.upenn.edu}.
We acknowledge support from ARL DCIST CRA W911NF-
17-2-0181, NSF Grant CCR-2112665, and the NSF Graduate Research Fellowship. }}   

\begin{document}

\maketitle

\begin{abstract}
\changed{
As robots become increasingly capable, users will want to describe high-level missions and have robots infer the relevant details.
Because pre-built maps are difficult to obtain in many realistic settings, accomplishing such missions will require the robot to map and plan online.
While many semantic planning methods operate online, they are typically designed for well specified missions such as object search or exploration.
Recently, Large Language Models (LLMs) have demonstrated powerful contextual reasoning abilities over a range of robotic tasks described in natural language.
However, existing LLM-enabled planners typically do not consider online planning or complex missions; rather, relevant subtasks and semantics are provided by a pre-built map or a user. 
We address these limitations via \texttt{SPINE}, an online planner for missions with incomplete mission specifications provided in natural language.
The planner uses an LLM to reason about subtasks implied by the mission specification and then realizes these subtasks in a receding horizon framework. 
Tasks are automatically validated for safety and refined online with new map observations.
We evaluate \texttt{SPINE} in simulation and real-world settings with missions that require multiple steps of semantic reasoning and exploration in cluttered outdoor environments of over 20,000m$^2$. 
Compared to baselines that use existing LLM-enabled planning approaches, our method is over twice as efficient in terms of time and distance, requires less user interactions, and does not require a full map. Additional resources are provided at \url{https://zacravichandran.github.io/SPINE}.
}
\end{abstract}



\glsresetall

\section{Introduction}
\label{sec:intro}

\begin{figure}[ht!]
    \centering
    \includegraphics[width=1\linewidth]{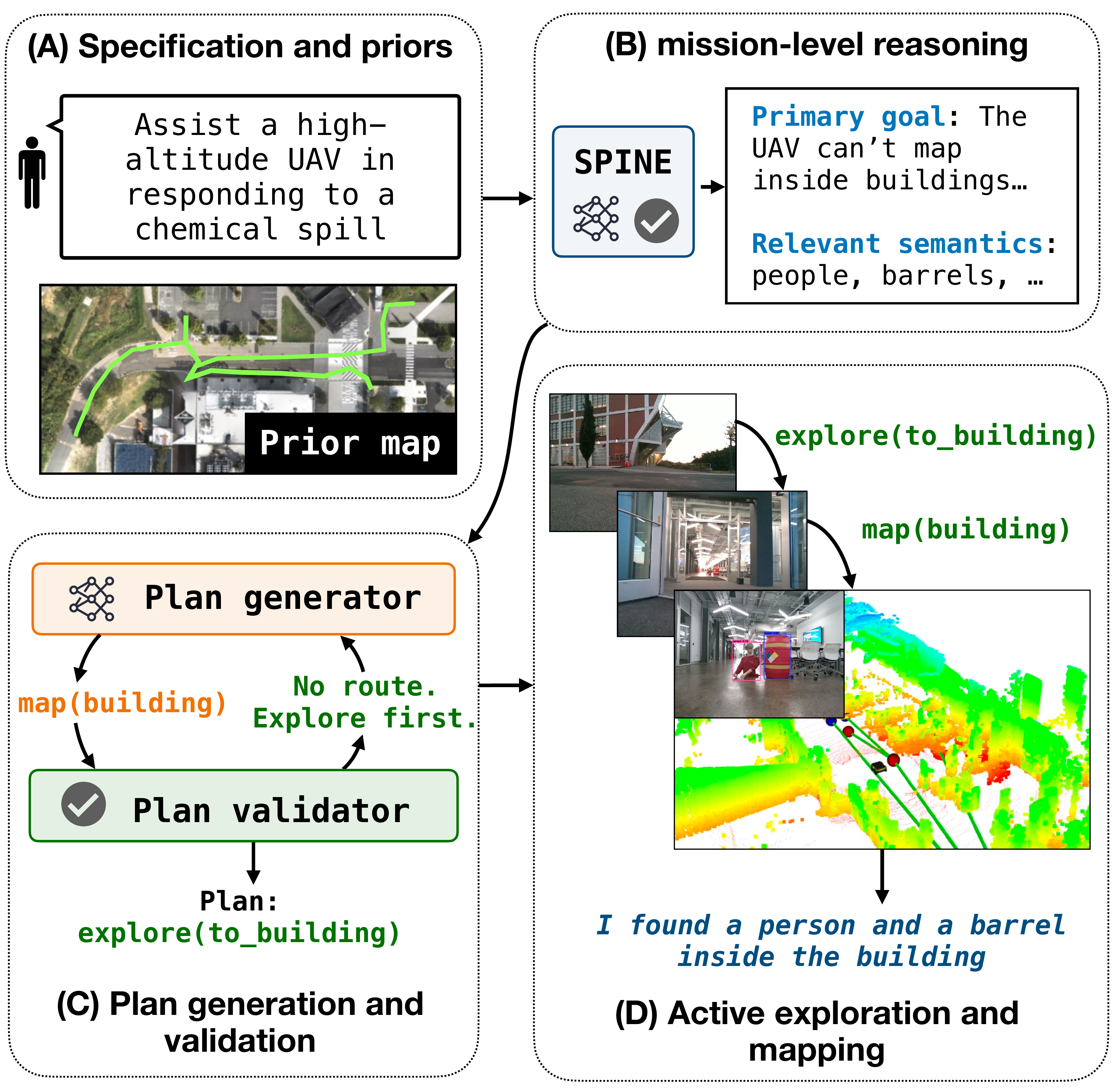}
    \vspace{-12pt}
    \caption{\changed{
    (A) \texttt{SPINE} takes as input a mission with incomplete specifications and prior map. (B) \texttt{SPINE} then reasons about the goals and semantics required to achieve the mission.
    (C) \texttt{SPINE}'s plan generator uses an LLM to generate subtasks, while its validation module ensures those subtasks are realizable. (D) \texttt{SPINE} then actively explores and reasons over acquired information in order to complete its mission.
    }
    }
    \label{fig:example_mission}
    \figvspace
    \vspace{-8pt}
\end{figure}

Consider an inspection robot operating after a heavy storm. 
A user may provide the following mission specification: \emph{Communications are down. Why?}
The robot will have to explore missing or changed regions of the map, locate relevant semantic entities (\textit{e.g.,} communication infrastructure), and collect precise mission-relevant information to assess infrastructure damage.
We refer to these mission specifications as \textit{\taskname} because
they imply subtasks and semantic targets that are not directly given to the robot; rather, they must be inferred from context. 
Because such missions often occur in partially-known environments, an autonomous robot must actively map its surroundings and refine its plan online.

Semantic planning methods have made progress on tasks such as object search, inspection, exploration, and mobile manipulation~\cite{active_3d_sem_slam, Kim2021PLGRIMHV, Ginting2024rss, spomp, canopy, Ginting2023icra,vasilopoulos_ral_2020, yokoyama2024vlfm, mobile_manipulation_icra_2022}. 
These methods typically maintain a semantic map of the environment such as a metric-semantic grid, object-oriented map, or scene graph, which the planner reasons over in search of its goal~\cite{ray2024taskmotionplanninghierarchical, vasilopoulos_ral_2020, spomp}.
With advances in semantic mapping, these representations can be built in real-time, which enables online planning~\cite{hughes2024foundations,liu2024slideslam, voxbloxpp,outdoor_dsg},
and some of these approaches are robust enough to be fielded in large scale environments~\cite{spomp_journal, chang2022lamp20robustmultirobot,kurtz2024realworlddeploymenthierarchicaluncertaintyaware}.
However, online semantic planners typically require a well-specified mission (\eg \hspace{0.5pt} \emph{inspect all the antennas in Zone A}).
And while formal  planning languages enable complex mission-level specifications~\cite{gosrich2023multitask, kalluraya2023multirobotmissionplanningdynamic, verginis2022planningcontrolmultirobotobjectsystems},
they still require a human operator to explicitly compose subtasks.

Recent work has addressed these limitations by using Large Language Models (LLMs) -- which have demonstrated powerful contextual reasoning over many domains -- to plan over tasks described in natural language~\cite{omama2023altpilotautonomousnavigationlanguage, wang2023voyageropenendedembodiedagent, codeaspolicies2022, chen2022nlmapsaycan}. 
Researchers have applied LLM-enabled planners to problems including mobile manipulation, navigation, and fault detection~\cite{codeaspolicies2022, saycan2022arxiv, rana2023sayplan, SinhaElhafsiEtAl2024Aesop, tagliabue2023real,vlmaps, huang2022inner}.
However, LLM-enabled planners typically require a pre-built map~\cite{rana2023sayplan, saycan2022arxiv, vlmaps, chen2023nl2tl, dai2024optimalscenegraphplanning}, and these methods generally consider well-specified missions~\cite{liu23lang2ltl, verify_llm_ltl, chen2023autotamp, codeaspolicies2022, vlmaps}. 
These assumptions prevent current LLM-enabled planning methods from operating in partially-known and unstructured environments, such as large-scale outdoor settings.

To address these limitations, we present \texttt{SPINE}, an online semantic planner for missions with incomplete specifications described in natural language. As shown in Fig.~\ref{fig:example_mission}, \texttt{SPINE} receives as input a specification and incomplete prior map.
\texttt{SPINE} then uses an LLM to reason about mission-level goals and relevant semantics, and then infers appropriate subtasks comprising navigation, active mapping, and user interaction.
These LLM-generated subtasks are validated for physical safety and syntactic correctness, which prevents unsafe actions from being executed by the robot.
Validated subtasks are realized in a receding horizon manner and are refined online as \texttt{SPINE} actively builds a semantic map.
By leveraging external data sources such as satellite imagery or UAV-generated maps, \texttt{SPINE} can operate in partially-known and unstructured environments.
To summarize, the contributions of the paper are:
\begin{enumerate}
    \item [\textbf{1.}]  An online semantic planner for language-specified missions in partially-known, unstructured  environments. 
    \item [\textbf{2.}] A plan generation module to infer subtasks from incomplete specifications and refine the subtasks online.
    \item [\textbf{3.}] A verification module that enables an LLM to safely propose navigation and exploration goals in unstructured and partially-known environments.
\end{enumerate}

We evaluate our method in large-scale outdoor simulated and real-world environments with missions that include semantic route inspection, multi-object search, and air-ground teaming.
Compared to LLM-enabled planning baselines that use either fully-known prior maps or receive explicit mission specifications, \texttt{SPINE} achieves comparable mission success while requiring significantly less time and user input.

\begin{figure}[th!]
    \centering
    \includegraphics[width=0.99\linewidth]{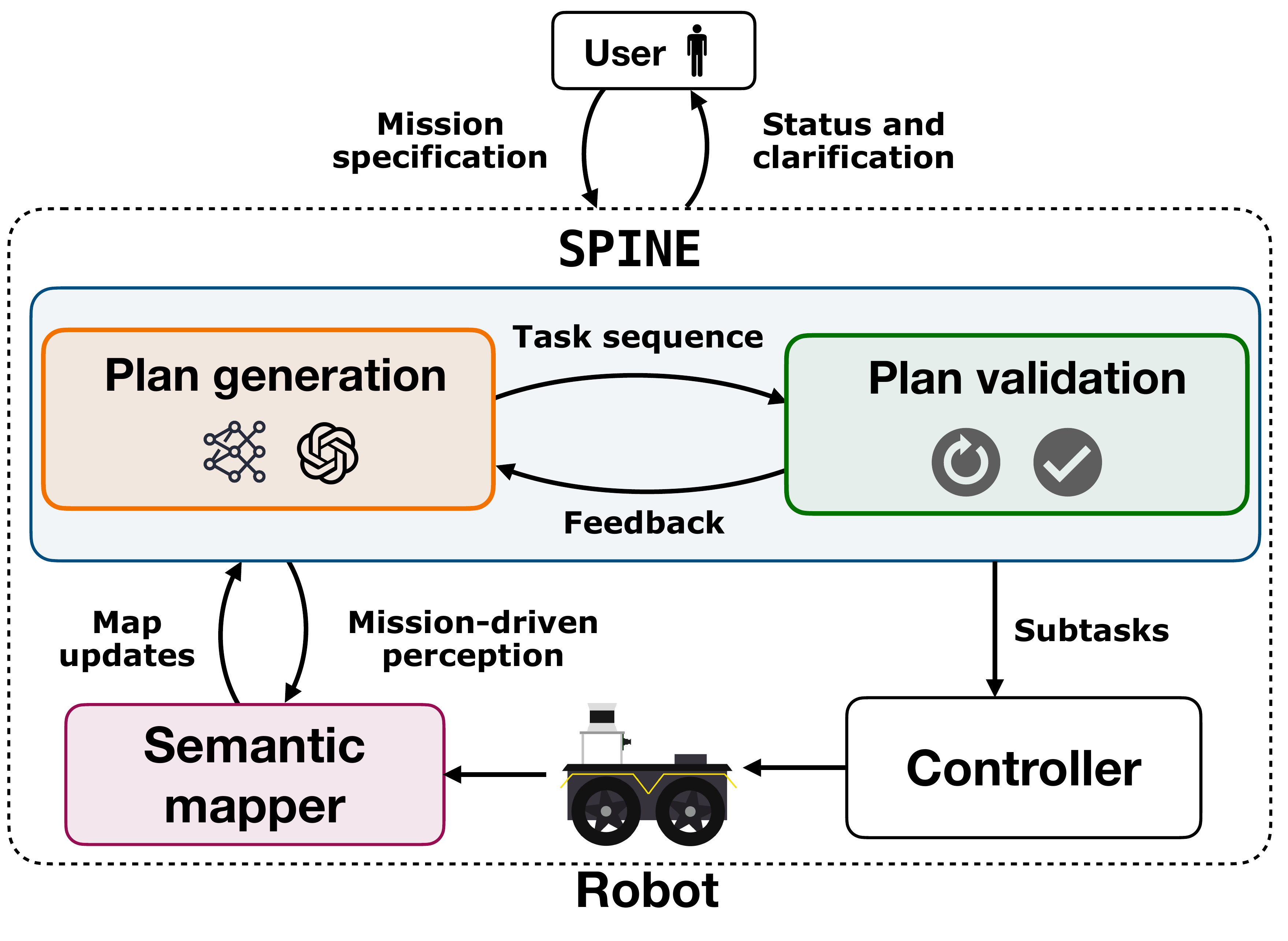}
    \vspace{-18pt}
    \caption{\changed{
    \texttt{SPINE} architecture. A user provides \texttt{SPINE} with a mission specification. 
    \texttt{SPINE} plans via behaviors for user interaction, active mapping, and robot control. 
    \texttt{SPINE}'s plan generator infers a task sequence which is validated online for correctness and feasibility; if necessary, feedback and corrections are provided in real-time. Actions are sent to the appropriate module, and the planner refines its plan as new information is acquired.
    }}
    \label{fig:sys-overview}
    \figvspace
\end{figure}

\section{Related Work}

\textbf{Representations for Semantic  Planning.} Effective planning  representations capture traversability, semantics, and spatial relationships needed for reasoning over contextual goals.
Advances in semantic mapping have enabled online planning for  tasks such as active exploration or object search ~\cite{Asgharivaskasi-TRO23, active_3d_sem_slam, spomp, spomp_journal, kurtz2024realworlddeploymenthierarchicaluncertaintyaware}. 
Scene graphs are a common representation for semantic planning, as they concisely represent objects, topology, and traversable regions~\cite{hughes2024foundations, outdoor_dsg}.
Semantic topological maps also provide a graphical scene representation but do not include a hierarchy~\cite{RoboHop, chiang2024mobilityvlamultimodalinstruction}.
Recent work incorporates foundation models into mapping pipelines in order to create open-vocabulary representations.
For example, ConceptGraphs~\cite{conceptgraphs}, HOVSG~\cite{werby23hovsg}, and Clio~\cite{Maggio2024Clio} assign semantic feature vectors to entities in the map, then task-relevant labels are assigned at runtime. 
\texttt{SPINE} is compatible with such state of the art mapping methods. In our experiments, we use an open-vocabulary semantic-topological mapper, which allows the planner to configure mission-specific semantics at runtime and operate in unstructured outdoor environments. 

\shortskip
\noindent\textbf{Online Semantic Planning.} Semantic planners reason over objects, regions, or other contextual information to solve tasks such as object search, inspection, and semantic exploration~\cite{active_3d_sem_slam, Ginting2024rss, Ginting2023icra, canopy, semantic_trajectory}.
Many works address online planning in partially-known environments, the planner's initial action sequence is informed by priors and refined online with new observations~\cite{Ginting2024rss, Ginting2024rss, vasilopoulos_ral_2020}.
Beyond object-level reasoning, semantic information also accelerates exploration of partially-known or unknown environments~\cite{zhou2023esc, active_3d_sem_slam, yu2023frontier}.
Fusing semantic knowledge from foundation models with classical search methods such as frontier exploration has been shown to an especially effective exploration strategy~\cite{yokoyama2024vlfm, zhou2023esc}.
Structured or formal planning languages, such as Linear Temporal Logic (LTL), may be used to compose more complex missions~\cite{kalluraya2023multirobotmissionplanningdynamic, verginis2022planningcontrolmultirobotobjectsystems,ray2024taskmotionplanninghierarchical, pacheck2023physically, menghi2019specificationpatternsroboticmissions}.
Notably, these methods require detailed mission specifications from a user, whereas our method infers mission details.

\shortskip
\noindent\textbf{LLMs for Planning.} Language has emerged as a powerful representation for specifying tasks, and
LLM-enabled planners have been applied to domains including mobile manipulation~\cite{rana2023sayplan, saycan2022arxiv, momallm24}, service robotics~\cite{llm_service_robot}, autonomous driving~\cite{sharan2023llm}, navigation~\cite{xie2023reasoning, pmlr-v229-shah23c, pmlr-v205-shah23b, vlmaps}, and fault detection~\cite{SinhaElhafsiEtAl2024Aesop, tagliabue2023real}
These methods typically configure an LLM via in-context or system prompts
with a problem description and a set of action primitives such as graph navigation goals~\cite{sharan2023llm}, lower-level application programming interface (API) for code generation~\cite{codeaspolicies2022, ma2023eureka, ma2024dreureka}, or learned behaviors~\cite{saycan2022arxiv}.
At runtime, the LLM is given an task specification and a map, such as a graph~\cite{rana2023sayplan} or semantic regions~\cite{huang2022inner}, then produces an action sequence.
A line of research develops LLM-enabled planners that translate mission specifications to a formal language such as Linear Temporal Logic (LTL) or Planning Domain Definition Language (PDDL)
~\cite{dai2024optimalscenegraphplanning, liu23lang2ltl, pmlr-v229-liu23d, chen2023nl2tl, chen2023autotamp, garg2024largelanguagemodelsrescue, liu2023llmp}.
While these instructions are complex, they explicitly state subtasks and semantic referents~\cite{verify_llm_ltl}.
Other research relaxes the requirement of a pre-built semantic map by incorporating feedback from perception systems~\cite{huang2022inner, momallm24, SinhaElhafsiEtAl2024Aesop} or specifying semantics at runtime~\cite{verify_llm_ltl, chen2022nlmapsaycan}. 
However, perception is limited to object detection or designed for small room-centric environments where the planner can leverage clear hierarchy and natural bounds on the environment.
In contrast, \texttt{SPINE} reasons over under-specified missions, does not require a pre-built map, and can operate in large unstructured environments. 

\section{SPINE} 
\subsection{Problem Statement}
\label{sec:problem_statement}
\changed{
We consider a robot that operates in an unstructured environment 
and is capable of performing behaviors such point navigation or area exploration. 
The robot's planner is provided with a mission specification in natural language, $S$.
Importantly, this specification is \textit{incomplete}, meaning it implies a goal and corresponding action sequences (\textit{i.e.,} subtasks) which are \textit{unknown} to the planner, thus planner must infer an action sequence that fulfills that goal with minimal planning  iterations.
The planner is also provided with a map, $M_k$ which is updated at each planning iteration, $k$, by an onboard mapper, and its previous actions, $a_{1:k}$.
At each iteration, the planner provides an action sequence, $\pi(S, M_k, a_{1:k}) \rightarrow a_{k+1:H}$, where $H$ denotes a planning horizon greater than $k$. This plan is realized in a receding horizon manner and refined online. 
}

\begin{table*}[ht!]
\vspace{3mm}
    \begin{tabular}{p{2.5cm}ccp{4cm}c} \toprule
         Purpose & Function & Arguments & Behavior & Constraints \\ \toprule
       \multirow{3}{2.5cm}{Navigation}  & \verb |map_region| & region node & navigate to goal and find objects & syntax, reachable\\
         &  \verb |exlore_region| & goal region, exploration radius  & explore around goal & syntax, reachable, explorable \\ 
           &  \verb |extend_map| & 2D coordinate & add frontier at coordinate & syntax, explorable \\
         & \verb |goto| & region & navigate to region & syntax, reachable \\ \midrule
       \multirow{2}{*}{\centering Active Mapping}           & \verb |inspect| & object and query & Inspect object & syntax  \\
         &  \verb |set_labels| & list of labels & Configure object detection & syntax \\ \midrule
        \multirow{2}{*}{User interaction} &  \verb |clarify| & question & ask for clarification from user & syntax \\
        &  \verb |answer| & provides answer & denotes task is complete & syntax \\
        \toprule
    \end{tabular}
    \caption{Available behaviors used by the semantic planner to compose action sequences.}
    \label{tab:actions}
    \tabvspace
    \vspace{-5mm}
\end{table*}

\subsection{\texttt{SPINE} Overview}
\changed{
\texttt{SPINE} comprises two modules -- a \emph{plan generator} and \emph{plan validator} -- as outlined in Fig.~\ref{fig:sys-overview}.
Along with the mission specification, online map, and action history, the plan generator receives online feedback, $f$, and proposes candidate actions $\pi_g(S, M_k, a_{1:k}, f) \rightarrow a_{k+1:H'}'$ (\S\ref{sec:llm-planner}).
The plan validator ensures that these actions are syntactically correct and physically realizable given the current map,  $\pi_v(a'_{k+1:H}, M_k) \rightarrow (f, a_{k+1:H'})$, where and $a_{k+1:H}$ is a validated action sequence and $f$ denotes feedback that may be used by the generator to correct erroneous action sequence (\S\ref{sec:verification}).
We outline \texttt{SPINE}'s inference process in Alg.~\ref{alg:verification2}.
}

\subsection{Semantic Mapper}
\label{sec:mapper}

Our architecture assumes a topological graph-based semantic mapper,
where nodes are of type \textit{region} or \textit{object}.
Regions indicate traversable points in freespace, and objects represent semantic entities. 
Edges in the graph are defined between either two regions (``region edges'') or an object and a region (``object edges'').
Regions edges denote paths traversable by the robot, while object edges denote that an object is visible from a certain region. 
Nodes may be enriched with additional semantic information (\eg ``this region is in a busy parking lot'', ``this car is damaged''), which provides \texttt{SPINE} with additional cues for planning. 
The mapper also maintains a local occupancy map which \texttt{SPINE} uses for action validation (\S\ref{sec:verification}).
The mapper is initialized with priors from satellite imagery, UAV maps, or previous mission data and is updated at each iteration.

\subsection{Plan Generator}
\label{sec:llm-planner}
The plan generator must infer action sequences that best fulfill the incomplete mission specification, which may require exploring previously unknown portions of the environment.
We instantiate the plan generator with an LLM, given their contextual reasoning abilities. 
We configure the LLM via a system prompt that describes its role, a definition of the mapping interface, and a description of the robot's behaviors.
At each iteration, the plan generator's four inputs -- the specification $S$, map $M_k$, action sequence $a_{1:k}$, and feedback $f$ -- are serialized into a textual representation and provided to the LLM via in-context prompts.

\shortskip
\noindent\textbf{Textual mapping interface.} The plan generator receive the prior semantic map in the following JSON schema:
\begin{tcolorbox}[colback=gray!3, colframe=black,left=1mm, right=1.5mm, top=1.5mm, bottom=1mm] \footnotesize
\begin{minted}[]{json}
{"regions": [{"name": "node_name", 
             "coordinates": "..."}, "..."],
"objects": ["..."],
"region_edges": [["source", "target"], "..."],
"object_edges": ["..."]}
\end{minted}
\end{tcolorbox}
\noindent 
Nodes are defined as a dictionary of attributes; this dictionary must contain the node's \verb|name| and \verb|coordinates|, but may be enriched with additional information such as a response to a mission-specify query (example in Fig.~\ref{fig:sim-radio-tower}).
Edges are simply defined by tuples of source and target nodes.
At each planning iteration, all map updates are provided to the plan generator via  in-context prompts which utilize the following API for high-level graph manipulation: \verb|add_nodes|, \verb|remove_nodes|, \verb|add_edges|, \verb|remove_edges|, \verb|update_nodes|.

\shortskip
\noindent\textbf{Reasoning for planning.} 
The plan generator provides an action sequence drawn from a predefined library of atomic behaviors (Alg.~\ref{alg:verification2} line 2). Our implementation uses behaviors for navigation, active perception, and user interaction (Tab.~\ref{tab:actions}), though in general these may include any feasible robot action. 
The generator parameterizes these behaviors with arguments that refer to the current semantic map or mission specification, and this process employs chain-of-thought (CoT) reasoning which explicitly states \texttt{SPINE}'s \emph{primary goal}, the \emph{relevant semantic graph} for the  mission, and a \emph{justification} for the proposed action sequence \cite{cot_llm}. 
During the first planning iteration, generator also provides \emph{relevant semantics} which may be used used to configure an open-vocabulary semantic mapping framework. 
We find that enforcing reasoning at multiple levels of abstraction helps the LLM to maintain a focus on its high-level goal while iteratively planning over shorter horizons. 
We illustrate one step of this process with a simplified example:
\medskip 

\noindent
\textbf{Example III.1}
Consider a robot provided with the mission specification \textit{I need to cross the river.} and a semantic map containing two regions -- \verb|region_1| located at \verb|(0, 0)| and \verb|dock| located at \verb|(0, 1)| -- and one edge that connects these two regions.
There is also a \verb|boat| located at \verb|(0, 2)| and connected to the \verb|dock|, but that boat is \textit{initially unknown}.
\texttt{SPINE} would receive the following map representation:
\begin{tcolorbox}[colback=gray!3, colframe=black,left=1mm, right=1.5mm, top=1.5mm, bottom=1mm] \footnotesize
\begin{minted}{json}
{"regions": [
  {"name": "region_1", "coordinates": [0, 0]}
  {"name": "dock", "coordinates": [0, 1]}],
"objects": [],
"region_edges": [["region_1", "dock"]],
"object_edges": []}
\end{minted}
\end{tcolorbox}

\noindent Because a boat would enable the user to cross the river, and boats are often found near docks, a reasonable output from the plan generator would be as follows:
\begin{tcolorbox}[colback=gray!3, colframe=black,left=1mm, right=1.5mm, top=1.5mm, bottom=1mm] \footnotesize
\begin{minted}{json}
{"primary goal": "find the user something 
   with which to cross the river",
"relevant semantics": ["boat"],
"relevant graph": "dock_1"
"justification": "boats are often found near 
   docks, so I should map that region",
"plan": "map_regin(dock_1), replan()"}
\end{minted}
\end{tcolorbox}
\noindent The robot's semantic mapper would be configured to look for \verb|boats|, and the robot would then map the \verb|dock|, upon which the semantic mapper would discover the boat and provide \texttt{SPINE} with the following update
\begin{tcolorbox}[colback=gray!3, colframe=black,left=1mm, right=1.5mm, top=1.5mm, bottom=1mm]\footnotesize
\begin{minted}{python}
add_node(boat, 
 attributes={"coordinates": [0, 2]}, 
 aedges=["dock"])
\end{minted}
\end{tcolorbox}
\noindent \texttt{SPINE} would then provide the user with the answer ``I found a boat near the dock that you could use to cross the river.''

\subsection{Plan validation} 
\label{sec:verification}
Valid action sequences require that all actions correctly invoke the behavior library while respecting constraints such as traversability; while LLMs provide powerful contextual reasoning abilities, they may hallucinate these details.
The plan validation module ensures that all actions satisfy pre-defined constraints before being realized by the robot (Alg.~\ref{alg:verification2} line 3). 
Any feedback provided by the validator is appended to the mission history (line 4), and \texttt{SPINE} returns its first valid plan sequence (line 5-6). 
If the plan generator exceeds a predefined iteration limit (\textsc{Max\_attempts}), the user is notified and invited to re-task \texttt{SPINE} (lines 7-8).

\begin{figure}[h!]
    \centering
    \includegraphics[width=0.99\linewidth]{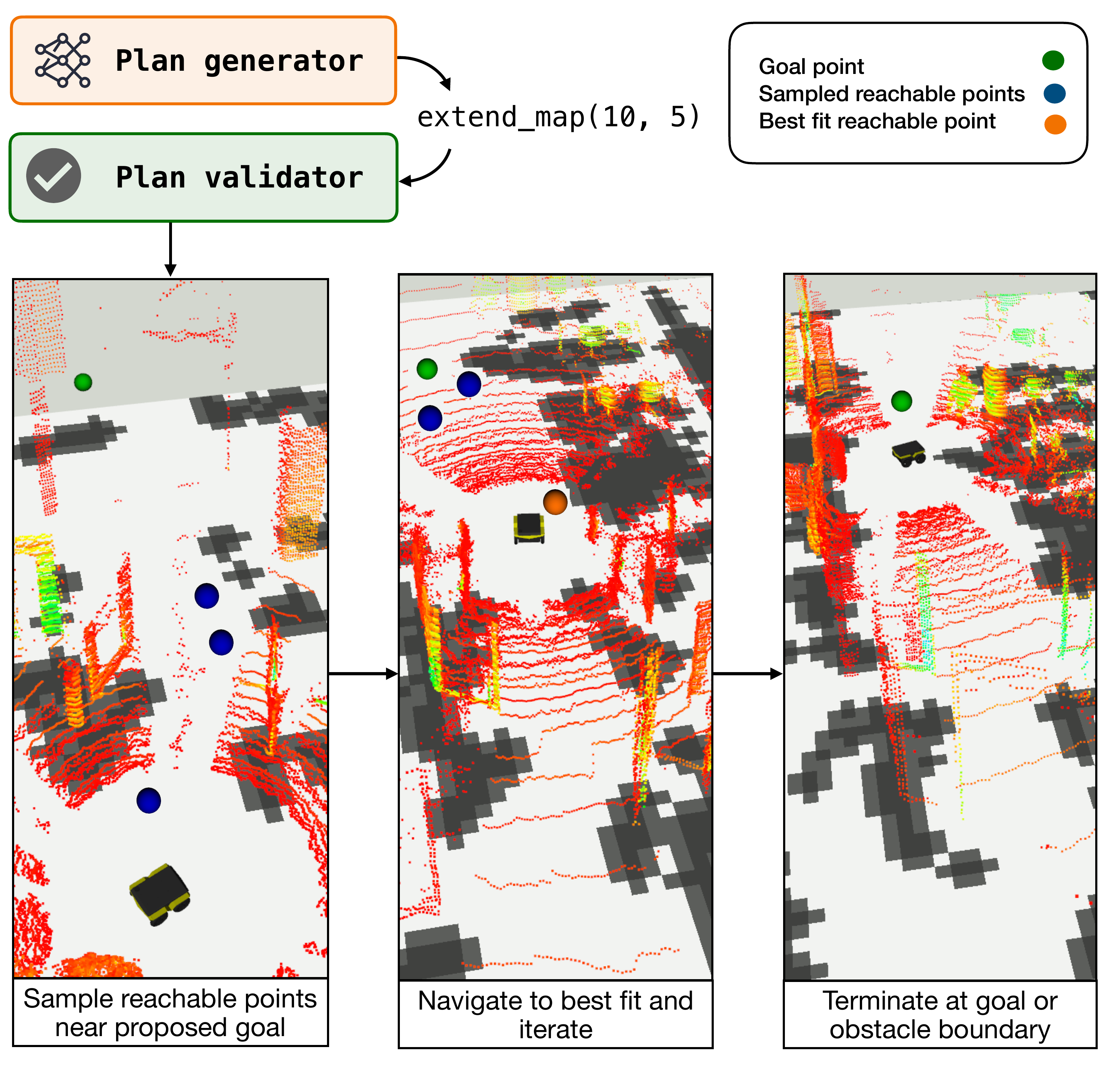}
    \vspace{-18pt}
    \caption{Online validation enables exploration. \texttt{SPINE}'s plan generator may produce exploration goal outside robot's obstacle map, which may not be reachable. Spatial validation iteratively finds best reachable fit, and the robot navigates to that point. Procedure terminates once robot reaches its goal}
    \label{fig:exploration_validation}
    \figvspace
\end{figure}

The validation module defines three types of constraints:  \emph{syntax}, \emph{reachability}, and \emph{explorable}.
Syntax constraints require that behaviors are invoked with the correct argument type.
Reachability constraints apply to navigation behaviors, and they require that a path to the navigation target exist \emph{within the current map}. 
These first two constraints are enforced upfront (lines 9-15); if violated, the validation module will provide specific feedback about the offending action, and the generator will produce a new task sequence.
\emph{Explorable} constraints are assigned to behaviors where the robot discovers new regions in the map.
We implement this constraint via frontier-style exploration to iteratively search for a traversable path towards a given goal.
The exploration terminates after reaching the goal or encountering an obstacle. 
For each breaking condition, semantic feedback is provided to the planner such as \verb|exploration terminated after|
\verb|encountering an obstacle| (lines 16-21, see Fig.~\ref{fig:exploration_validation}).

\begin{algorithm}[h]
\label{alg:verification2}
\small 
\SetKw{KwInit}{Initialize:}
\caption{\changed{SPINE inference procedure}}
\KwIn{\textsc{Specification} $S$, 
\textsc{Map} $M_k$, 

\textsc{Action history} $a_{1:k}$
\textsc{feedback} $f$}

\For {\textsc{Attempt} \normalfont{\textbf{in}} \textsc{Max\_attempts}}
{
$a'_{k+1:H'} \leftarrow \pi_g(S, M_k, a_{1:k}, f)$  \algorithmiccomment{generate actions}

($a_{k+1:H}, f_k) \leftarrow \pi_v(a'_{k+1:H'}, M_k)$ \algorithmiccomment{validate actions}

$f \leftarrow f + [f_k]$

\If{$a_{k+1:H} \neq \emptyset$}  
{
    \Return {$a_{k+1:H}$} \algorithmiccomment{return first valid sequence}
}
}

\Return \texttt{NotifyUser} \algorithmiccomment{if unable to create plan}

\vspace{6pt}
\textbf{function} $\pi_v(a'_{k+1:H'}, M_k)$ 

\Indp  
\textsc{Validated Actions} $a_k = []$

\textsc{Generator Feedback} $f = []$

\For {$a_i$ \normalfont{in} $a'_k$}
{
\If{\normalfont{\textbf{not}} \textsc{SyntactiallyValid}($a_i, M_k$)}
{
   $f \leftarrow$ \textsc{GetErrorFeedback}($a_i, M_k$)
   
   \Return $a_k, f$
}

\For {$a_i$ \normalfont{in} $a_{k+1:H'}'$}
{
    \If{\textbf{not} \textsc{explorable}($a_i$)}
    {
        $a_k \rightarrow a_k + [a_i]$
    } 
    $a_i, \text{result} \leftarrow$ \textsc{Explore}($a_i, M_k)$
    
    $a_k \leftarrow a_k + [a_i]$
    
    $f \leftarrow f + [\text{result}]$
}
\Return $a_k, f$
}
\end{algorithm}
\vspace{-2mm}

\section{Experiments}
\label{sec:experiments}

We design experiments to assess our contributions (\S\ref{sec:intro}):
\begin{enumerate}
    \item [\textbf{Q1}.]  Does \texttt{SPINE} provide  time and distance savings compared to offline LLM-enabled planning approaches? 
    \item [\textbf{Q2}.] Can \texttt{SPINE} achieve missions competitively compared to methods that are explicitly given a full prior map and mission specifications?
    \item [\textbf{Q3}.] How important is validation for online planning? 
\end{enumerate}

\noindent We use simulation and real robot experiments to answer \textbf{Q1} and \textbf{Q2}, and we design an ablation study to answer \textbf{Q3}.

\subsection{Implementation Details}

Both simulated and real robot experiments assume a mobile robot equipped with a Lidar and RGB-D camera. 
We implement the behaviors from \S\ref{sec:llm-planner} using ROS MoveBase~\cite{quigley2009ros}, and the plan generator uses the base GPT-4 model~\cite{openai2024gpt4technicalreport}. 
We implement a semantic mapper that provides the graph-base representation described in \S\ref{sec:mapper}.
The mapper uses Faster-LIO to estimate odometry~\cite{fasterlio} and GroundGrid~\cite{GroundGrid} to estimate free-space and establish region nodes.
GroundingDino~\cite{GroundGrid} provides open-vocabulary object detections, and these detections are associated using a multiple-hypothesis tracker. The mapper uses the LLaVA vision-language model to enrich the map with mission-relevant semantic descriptions~\cite{liu2023llava}.
All autonomy runs onboard, except for the LLM which makes API queries over an internet connection. 
We conduct simulation experiments in photorealistic Unity testbed, which provides sensor and control feeds for a ClearPath Husky. 
We then perform real robot experiments using a Clearpath Jackal equipped with a Ouster Lidar, Realsense RGB-D Camera, Nvidia RTX 4000 GPU, and Ryzen 5 3600 CPU.

\subsection{Experimental Setup}

\noindent\textbf{Baselines.}
We compare against two baselines: \textit{Explicit Tasking} and \textit{LLM-as-planner}.
\textit{Explicit Tasking} receives step-by-step instructions and resembles existing LLM-enabled planning methods where the user provides explicit mission instructions, such as those using formal methods,~\cite{liu23lang2ltl, pmlr-v229-liu23d}.
\textit{LLM-as-planner} 
resembles approaches which receive a full map upfront along with a mission specification~\cite{codeaspolicies2022,rana2023sayplan,saycan2022arxiv,verify_llm_ltl}; following previous work, this baseline can still discover new objects in the scene.~\cite{huang2022inner, saycan2022arxiv}.
\medskip

\noindent\textbf{Metrics.}
We report five metrics: \textit{mission success}, \textit{time} required for mission completion, \textit{distance} traveled for mission completion, LLM \textit{queries}, and user \textit{interactions} during the mission.
User interactions capture the complexity of the mission, as more complex missions will require more instructions from the user. 

\begin{figure}[t!]
\vspace{3mm}
    \centering
    \includegraphics[width=0.99\linewidth]{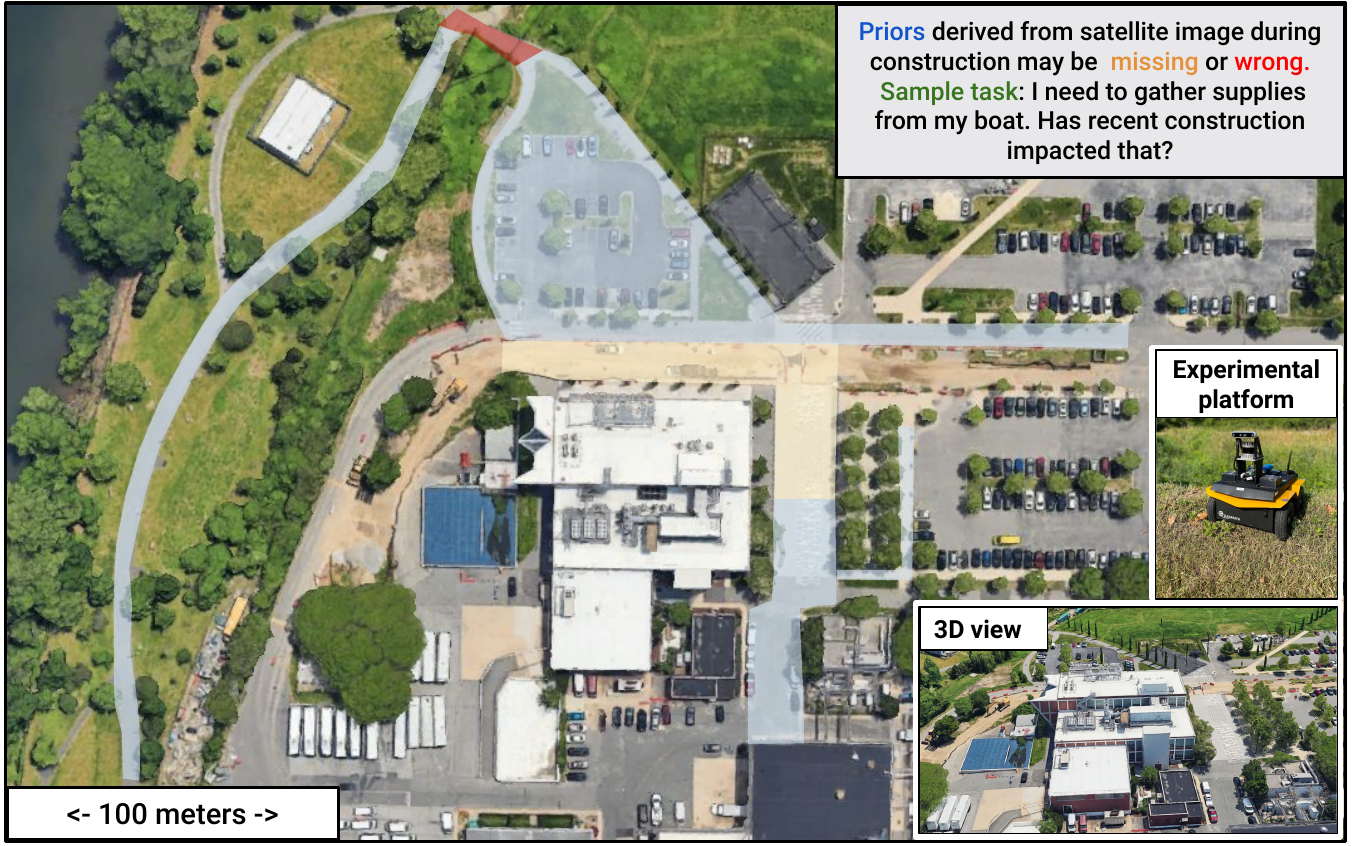}
    \vspace{-18pt}
    \caption{Experimental platform, 3D view of environment, and example prior and corresponding task used for real-world experiments. The prior map is derived from outdated satellite imagery or obstructed due to trees and other coverings. The prior map is thus incomplete and partially incorrect, which requires the planner to reason about information acquired online.}
    \label{fig:exp_setup}
    \vspace{-12pt}
\end{figure}

\begin{figure}[h!]
    \centering
    \includegraphics[width=0.95\linewidth]{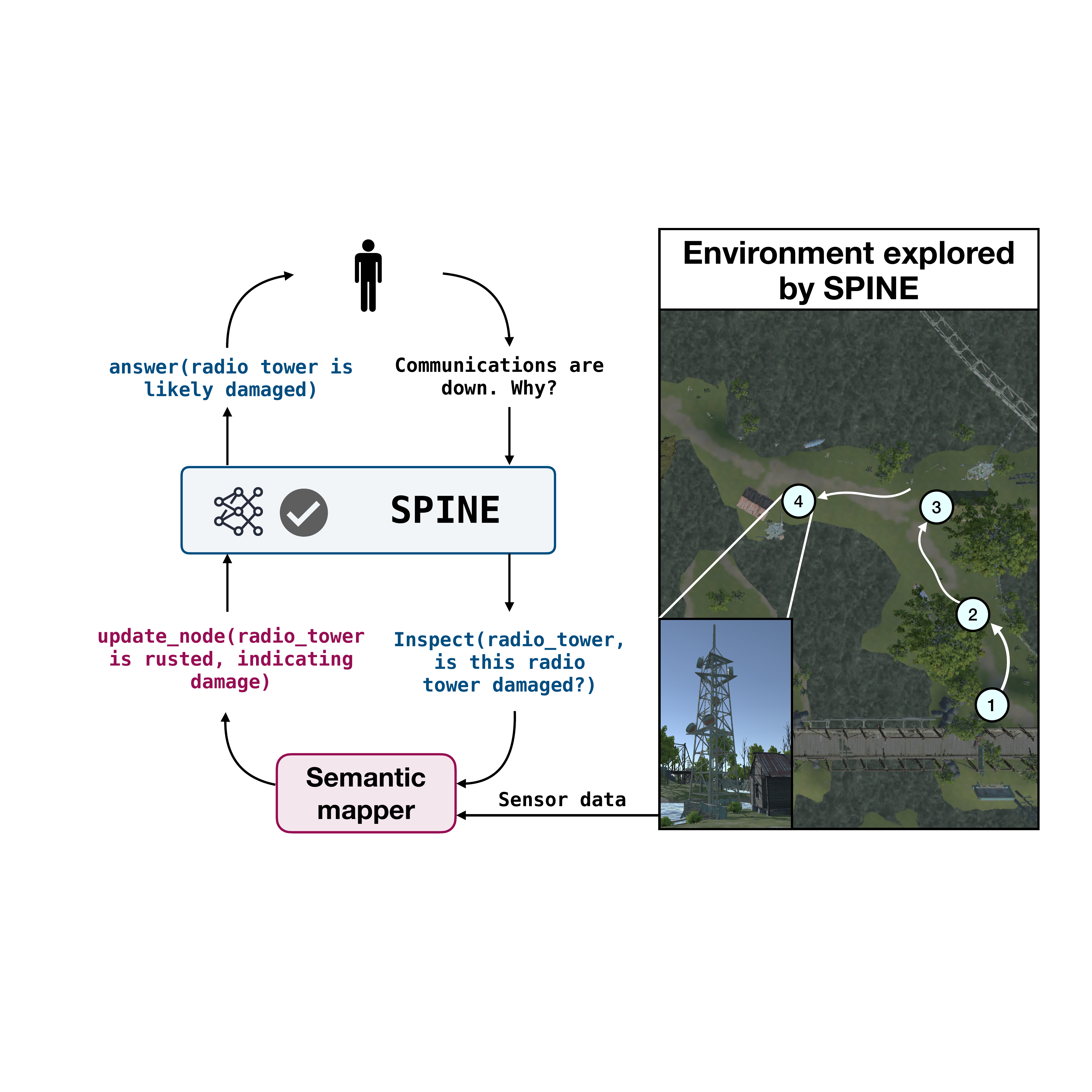}
    \caption{
    \changed{
    Given a mission and  prior map, \texttt{SPINE} must (2) explore and  (3, 4) visit and inspect communication infrastructure. \texttt{SPINE} then forms an appropriate inspection query for the mapper's vision language model (VLM), and it uses the acquired information to solve the mission.}
    }
    \label{fig:sim-radio-tower}
    \figvspace
    \vspace{-2pt}
\end{figure}

\medskip\noindent\textbf{Evaluation Environments.} 
We perform simulation experiments in a rural outdoor space of over 40,000m$^2$ where missions require the robot to travel over 400m (example in Fig~\ref{fig:sim-radio-tower}). 
Real experiments are conducted in an office park of over 20,000m$^2$ where missions require the robot to travel over 250m (Example in Fig~\ref{fig:exp_setup}).
We construct prior maps for both environments based on information that could be acquired from satellite imagery or UAV-generated maps.

\medskip\noindent \textbf{Mission Specifications.}
We consider missions with the following linguistic specifications:
\begin{enumerate}
    \item [1.] There was a storm last night. I am worried that impacted logistics, because I need to drop off supplies today. Can I still do that?
    \item [2.] I sent a robot out to collect supplies from an incoming boat. I have not heard back. What happened?
    \item [3.] Communications are down, why?
    \item [4.] I need to gather supplies from my boat. Has recent construction impacted that?
    \item [5.] You're assisting a UAV in response to a chemical spill. Triage regions not visible from the air.
\end{enumerate}
\noindent Each mission requires completing 2-8 subtasks of semantic reasoning and exploration.
We run each mission one to three times and vary the prior map and initial conditions.
See Fig.~\ref{fig:example_mission}, Fig.~\ref{fig:exp_setup}, and Fig.~\ref{fig:sim-radio-tower} for examples.

\subsection{Simulation Results}
As reported in Tab.~\ref{tab:agg-sim-results}, 
Explicit Tasking completes 100\% of missions while taking on average 532s, traveling 292m, making 8.6 API calls, and 4  user interacting.
Despite receiving only a partial map and incomplete specification, \texttt{SPINE} achieves a 94.3\% mission success while requiring similar time and distance. 
Notably, \texttt{SPINE} makes a similar number of LLM queries, which indicates it's iteratively inferring and realizing the subtasks given to Explicit Tasking (\textbf{Q2}).
Imperfect success rate comes from the third mission, where \texttt{SPINE} must inspect multiple communication towers for damage. After finding that the first tower is damaged, instead of inspecting the next tower, \texttt{SPINE} declares the mission complete. 
While the LLM-as-planner approach is competitive in terms of success, because it must fully map the environment it requires over twice the time and traversal distance required (\textbf{Q1}). 
This method does require less LLM queries as compared to \texttt{SPINE}; because it receives a full map, this planner can generate a near-complete task sequence at the first planning iteration.


\begin{table}[t!]
    \centering
    \caption{\changed{Simulation experiment results.  }}
    \label{tab:agg-sim-results}
    \adjustbox{max width=\columnwidth}{
    \begin{tabular}{ccccccc} \toprule
         \multirow{2}{*}{Method} & \multicolumn{5}{c}{Metrics} \\ \cmidrule(lr){2-6}
          &  Success  & Time & Distance  & Interactions & Queries \\ \midrule
        SPINE & 94.3\% & 536.6s  & 312.4m  &  1 & 6.6 \\ 
        LLM-as-planner &  100\% & 1244.7  & 677.4  & 1 & 1.7\\ 
        Explicit Tasking & 100\% & 523s  & 292m  &  4 & 8.6 \\    
   \bottomrule
    \end{tabular}}
    \tabvspace
\end{table}

\subsection{Real Robot Results}

As reported in Tab.~\ref{tab:real-exp}, Explicit Tasking takes 1035s, travels 202m, 8.6 API calls, and requires 5 user interactions to complete a mission on average.
\texttt{SPINE} still compares favorably to Explicit Tasking in terms of time, distance, and user interactions required (\textbf{Q2}).
Interestingly, \texttt{SPINE}'s mission success rate was higher than in simulation, which is likely due to the increased scale of the simulated environment. 
Also due to increased environmental scale, there is a comparatively larger gap between the  LLM-as-Planner approach and \texttt{SPINE} (\textbf{Q1}); SPINE requires less than one third of the time and 2.5 less distance to complete a mission. 
We noticed that average robot speed was slower across all methods, which was due to more complex perception input, greater actuation noise, and increased obstacles as compared to simulation.
See Fig.~\ref{fig:example_mission} for an example mission.

\begin{table}[!t]
    \centering
    \caption{ \changed{Real-world experiment results.}}
    \label{tab:real-exp}
     \adjustbox{max width=\columnwidth}{
    \begin{tabular}{ccccccc} \toprule
         \multirow{2}{*}{Method} & \multicolumn{5}{c}{Metrics} \\ \cmidrule(lr){2-6}
          &  Success  & Time & Distance  & Interactions & Queries \\ \midrule
        SPINE & 100\%  & 1126.0s & 224.0m & 1 & 8.3 \\ 
        LLM-as-planner & 100\%  & 3701.1s &  570.8m & 1 & 1.42 \\ 
        Explicit Tasking &100\%  & 1035.0s  & 202m   & 5  & 8.6 \\   
        \bottomrule
    \end{tabular}
    }
    \tabvspace
\end{table}

\subsection{Validation module ablation}
\begin{figure}[h!]
    \centering
    \includegraphics[width=0.95\linewidth]{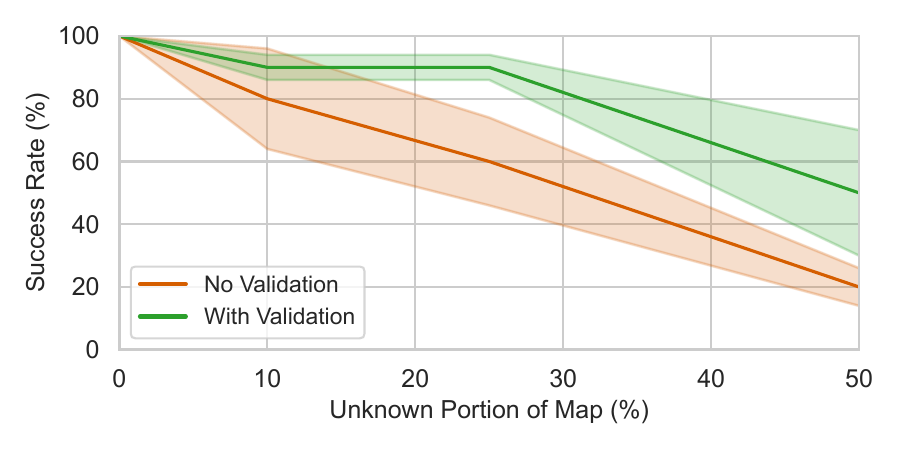}
    \vspace{-2mm}
    \caption{Validation ablation experiment results (mean and variance).}
    \label{fig:ablate}
    \vspace{0mm}
\end{figure}

We assess the importance of online validation by comparing our method to a variant without validation (\textbf{Q3}). 
We provide an identical specification to each method, and we measure mission success rate as we randomly remove portions of the prior map, averaged over four trials. 
Results, shown in Fig.~\ref{fig:ablate}, indicate that verification is increasingly important as the environment becomes less certain. 
Qualitatively, the LLM is prone to hallucinate connections and exploration goals. 
Validation prevents hallucinated goals from being realized on the robot and offers an alternative plan instead (See Fig.~\ref{fig:exploration_validation}).

\section{Conclusion}
We present \texttt{SPINE}, an online planner for missions with incomplete specifications in partially-known and unstructured environments.
\texttt{SPINE} uses an LLM to decompose specifications in natural language into a sequence of subtasks, comprising navigation, active mapping, and user interaction, which are automatically validated and refined online.
Simulation and real-world experiments demonstrate that
\texttt{SPINE} performs comparably to methods that receive step-by-step instructions from an expert user. 
\texttt{SPINE} is also more efficient in terms of distance and time required to complete a mission as compared to the two step process of first mapping and then using an LLM-enabled planner, and does not require full \textit{a priori} knowledge of the environment. 

Future work may take several directions. 
\texttt{SPINE} requires an internet connection for LLM queries, which requires a network infrastructure.
Going forward, we would like to mitigate this limitation by adapting smaller, open-sourced LLMs such as Llama~\cite{touvron2023llamaopenefficientfoundation} or Gemma~\cite{gemmateam2024gemmaopenmodelsbased} variants that are efficient enough to run onboard a robot.
And while this paper focuses on single-robot planning, we believe that extending \texttt{SPINE} for online and distributed multi-robot planning applications a natural extension this work.

\clearpage
\bibliographystyle{IEEEtran}
\bibliography{IEEEabrv, literature}

\clearpage

\appendices

\setcounter{figure}{0}
\setcounter{table}{0}
\renewcommand{\thefigure}{A\arabic{figure}}
\renewcommand{\thetable}{A\arabic{table}}
\renewcommand{\thesection}{A\arabic{section}}

\section{Summary}
In this appendix we provide further detail on our proposed method. 
Subsection~\ref{subsec:llm_config} provides details on the LLM system prompt, including the perception api and planning interface.
Subsection~\ref{subsec:behavior} describes the behavior library implementation including the controller used.
Subsection~\ref{subsec:perception} provides more details and visualizations on teh semantic mapping components traversability estimation, object detection, and VLM results.
Section~\ref{sec:experiment_details} provides details on the experimental setup. 
We provide more details on the experimental missions, including subtasks required and prior maps given to the planner 
and provide further discussion on results (Subsection~\ref{subsec:further_disussion}), including why the performance of SPINE was 6\% lower than baselines in the simulation experiments (see Tab.~\ref{tab:agg-sim-results})

\section{Further method details}
We provide details on the implementation of the LLM configuration, semantic mapper, and behavior library.

\renewcommand{\vm}[1]{{\color{teal}VM: #1}}
\subsection{LLM configuration}
\label{subsec:llm_config}
The LLM configuration consists of four main parts: main system configuration, perception API, planning API, and planning advice. 
The system configuration provides an overview of the LLM's role in the planning framework and defines interfaces (see Listing~\ref{listing:role}).
The perception API defines how the LLM will receive updates from the semantic mapper (see Listing~\ref{listing:perception}).
The planning API defines how the LLM will compose subtasks sequences (see Listing~\ref{listing:planning_api}).
Finally, the advice portion of the configuration preempts common mistakes we observed the LLM making during development (see Listing~\ref{listing:advice}).
We also provide five in-context examples of canonical planning behavior, and example of which is detailed in Listing~\ref{listing:in_context}, and we refer the reader to our software for a complete list.
At runtime, the user-provided mission and current scene graph is appended to the context.

\subsection{Behavior library and Constraint Feedback}
\label{subsec:behavior}
We provide further details on the behaviors listed in Tab.~\ref{tab:actions}.
\verb|goto| takes a string, which is interpreted as a region node. 
The planner with find the shortest path to that node over the current graph, and it will then navigate to that node.
The following behaviors call \verb|goto| for navigation to a particular node, where applicable.
\verb|map_region| takes a string, which is interpreted as a region node. The robot will navigate to that node and report any objects detected along the way.
\verb|explore_region| takes a string and float, which is interpreted as a region node and exploration radius, $r$, in meters. 
The robot will navigate to that node, then explore the circle of radius $r$ around that region node.
\verb|extend_map| takes two floats, which is interpreted a 2D coordinate. 
The robot will attempt to navigate to that coordinate.
\verb|inspect| takes two strings, which is interpreted as an object node and inspection query. 
The robot will navigate to that object, which is obtain an image of that object, pass that image and query to a VLM, and report the VLM answer.
\verb|set_labels| takes a set of strings, which is  interpreted as class labels. These labels are used to configure the object detector.
\verb|clarify| takes a string, interpreted as a question and provided to the user. The user can respond.
\verb|answer| takes a string, which is interpreted as an answer to the user's mission.
This terminates the mission.
For all navigation behaviors, we use the controller implemented by ROS Move Base~\footnote{\href{http://wiki.ros.org/move_base}{http://wiki.ros.org/move\_base}} with a target velocity of $0.5m/s$.

Each constraint provides tailored feedback, if violated. 

\medskip

\noindent\textbf{Syntax} is defined over the previously described behaviors. 
The feedback associated with this constraint highlights offending variables and function spelling.

\medskip

\noindent\textbf{Reachable} is defined over region nodes. There must be a path to the region node in the current map.
Feedback associated with this constraint lists unreachable nodes. 
Feedback will then suggest exploration objectives based on the closest reachable node to the goal point.

\medskip

\noindent\textbf{Explorable} is defined over exploration goals. There must be a obstacle-free path between the robot's current location and the goal.
If such a path cannot be found, feedback will provide the reason why (\eg, exploration hit an obstacle boundary).

\subsection{Semantic Mapper}
\label{subsec:perception}
\begin{figure*}
    \centering
    \includegraphics[width=0.99\linewidth]{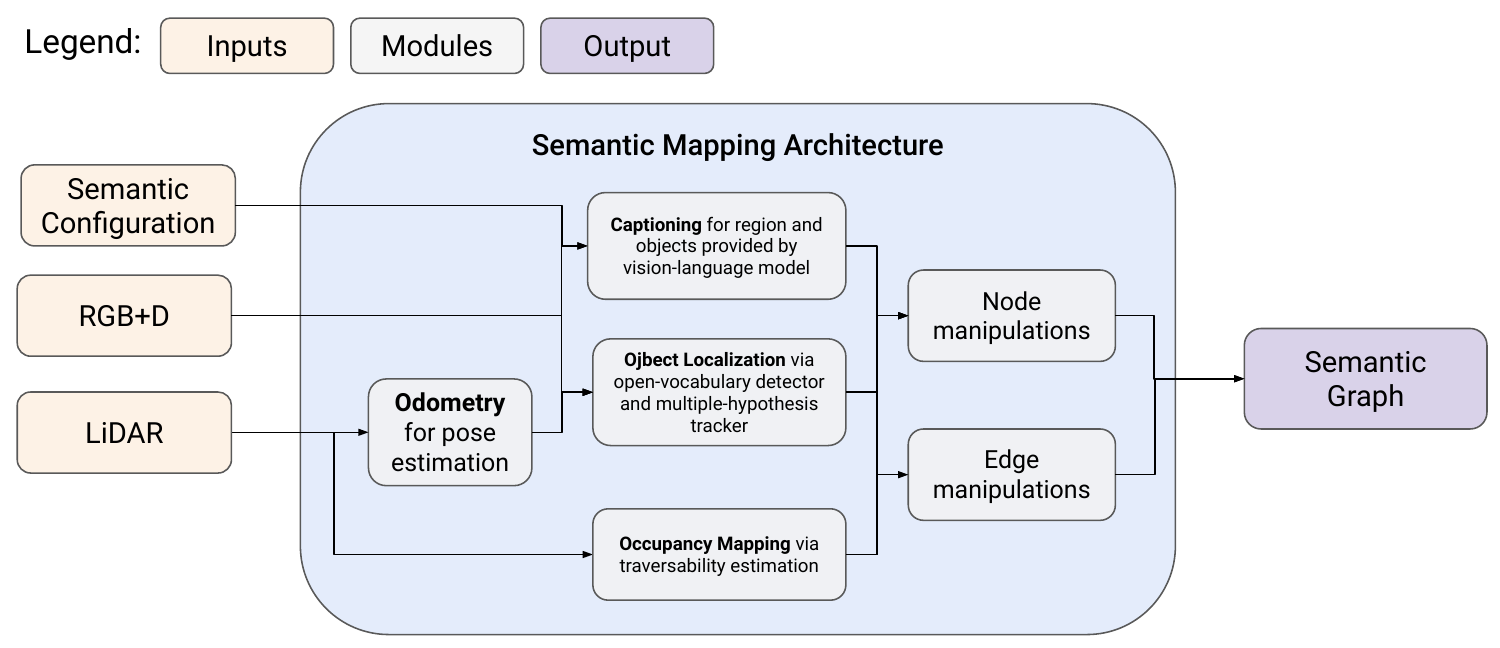}
    \caption{Semantic mapping architecture used by SPINE. The mapper takes LiDAR, RGB + Depth (RGB+D) sensor streams, and semantic configuration provided by the semantic planner. Odometry provides pose estimation. Occupancy mapping uses a traversability estimator to build a local map of obstacles, which is used to add and remove regions or edges to the map. Object localization uses an open-vocabulary object detector and multiple-hypothesis tracker to identify and ground objects in physical space. The captioning module provides further semantic detail to detected objects or regions. Information from these modules is uses to added and remove nodes and edges. Semantic configuration is used to set labels for the Object Localization module or provide queries for the Captioning module.}
    \label{fig:supp-perception}
\end{figure*}

\begin{figure*}
    \centering
    \includegraphics[width=0.99\linewidth]{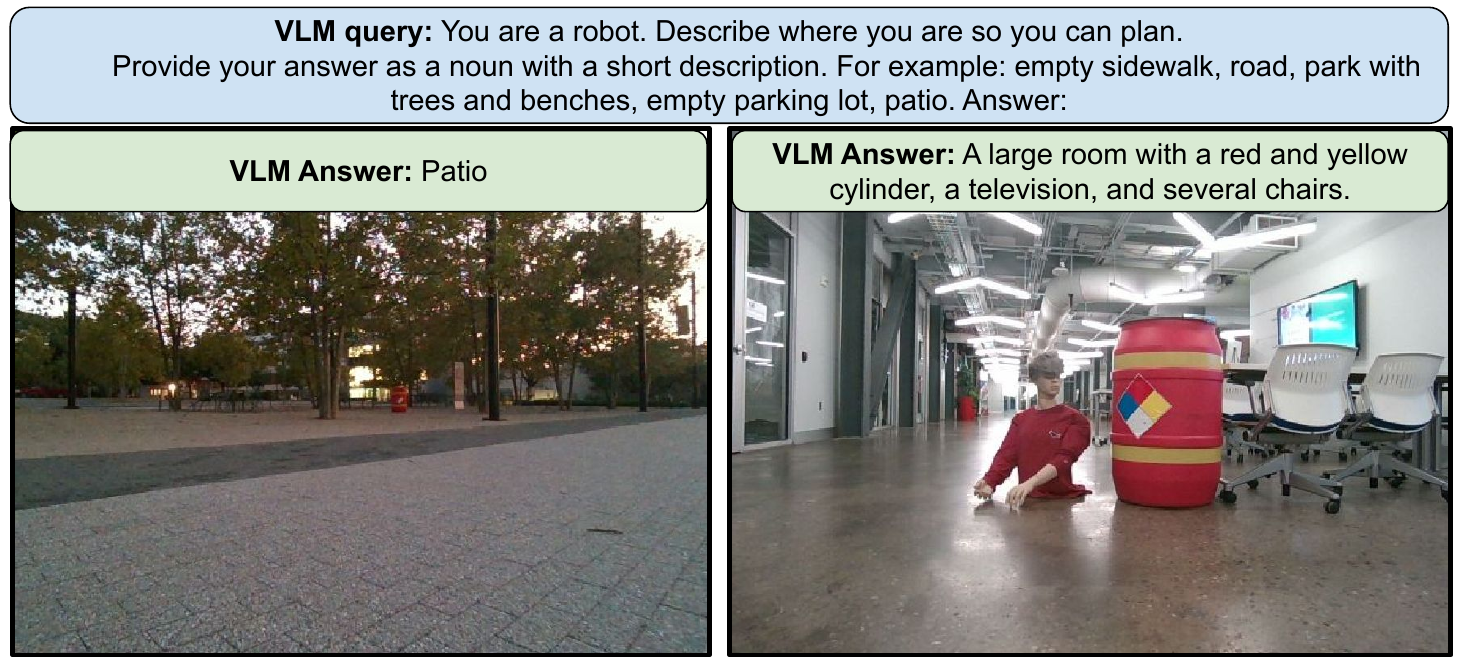}
    \caption{Examples of Vision-Language Model captioning during exploration. Captions provide brief semantic descriptions of the scene which may be useful for planning. }
    \label{fig:supp-vlm-region-1}
\end{figure*}

The architecture for the semantic mapper used by SPINE is shown in Fig.~\ref{fig:supp-perception}.
The mapper takes RGB + Depth, LiDAR, and semantic configuration as inputs.
LiDAR is used for odometry estimation (Faser-LIO~\cite{fasterlio}) and local occupancy map construction (GroundGrid~\cite{GroundGrid}).
The occupancy map is used to add and remove regions and edges from the map based on connectivity. 
RGB+D is used for object localization and captioning.
Objects are detected using GroundingDino~\cite{liu2023groundingdino}).
Detections are then clustered and localized with a multiple-hypothesis tracker.
A vision-language model (LLaVA~\cite{liu2023llava}) provides enriches the semantic information available to the planner (see Fig.~\ref{fig:supp-vlm-region-1}, Fig.~\ref{fig:sim-radio-tower}).
Outputs from these modules are used to add and remove nodes and enrich them with semantic information.
Semantic configuration is provided by the planner and is used to set the labels of the object detector and provide queries to the vision language model. 
The detection and tracking modules runs at roughly 5Hz, and the vision-language model runs at roughly 1Hz, and occupancy map construction runs well over 10Hz, all onboard.
Taken together, the semantic mapper runs sufficiently fast for real-time planning and control. 

\begin{figure*}
    \centering
    \includegraphics[width=0.99\linewidth]{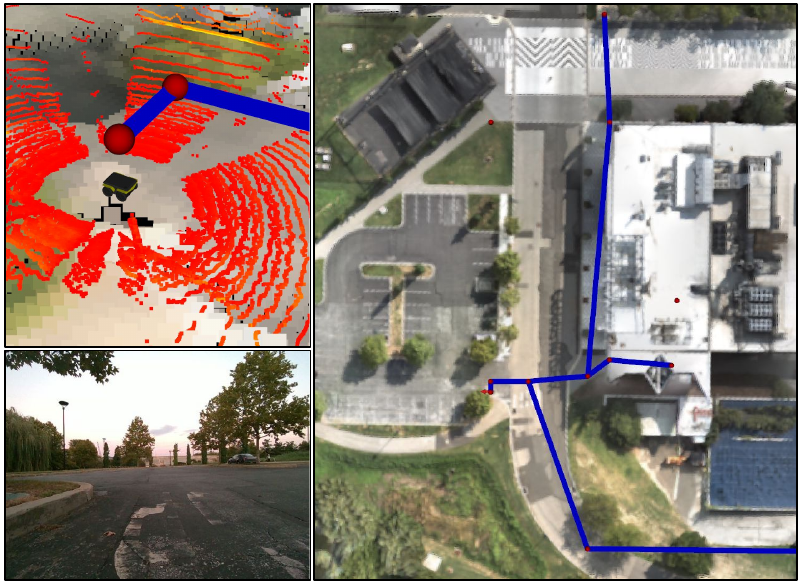}
    \caption{Example prior graph used by SPINE (right). Edges in blue and nodes in red. Semantic labels omitted for clarity. Third person view of robot is overlaid on overhead imagery (top left). Camera view from the robot is shown in the bottom left. Because this graph was derived from overhead imagery, registration was imperfect, and the planner must adjust in real-time (note edges that cross the building intersection).}
    \label{fig:supp-graph}
\end{figure*}

\clearpage

\begin{listing*}[h!]
\label{listing:role}
\caption{LLM system prompt: role description}
\begin{tcolorbox}[left=1mm, right=1.5mm, top=1.5mm, bottom=1mm] 
\footnotesize
\begin{minted}[]{text}
Agent Role: You are an excellent graph planner. You must fulfill a given task provided by 
the user given an incomplete graph representation of an environment.

You will generate a step-by-step plan that a robot can follow to solve a given task. 
You are only allowed to use the defined API and nodes observed in the scene graph for planning.
Your plan will provide a list of actions, which will be realized in a receding-horizon manner. 
At each step, only the first action in the plan will be executed. 
You will then receive updates, and you have the opportunity to replan. 
Updates may include discovered objects or new regions in the scene graph.
The graph may be missing objects and connections, so some tasks may require you to explore. 
Exploration means mapping existing regions to find objects, or adding a new region to find paths. 

The graph is given the in the following json format:
```
{
        "objects": [{"name": "object_1_name", 
        "coords": [west_east_coordinate, south_north_coordinate]}, ...], 
        "regions": [{"name": "region_1_name", 
        "coords": [west_east_coordinate, south_north_coordinate]}, ...],
        "object_connections: [["object_name", "region_name"], ...],
        "region_connections": [["some_region_name", "other_region_name"], ...]
        "robot_location": "region_of_robot_location
}
```

Each entry of the graph contains the following types:
- "regions" is a list of spatial regions. 
The regions are traversable ONLY IF they appear in the "region_connections" list
- "object_connections" is a list of edges connecting objects to regions in the graph. 
An edge between an object and a region implies that the robot can see 
the given object from the given region
- "region_connections" is list of edges connecting regions in the graph. 
An edge between two regions implies that the robot can traverse between those regions.


Provide you plan as a valid JSON string (it will be parsed by the `json.loads` function in python): 
```
{
"primary_goal": "Explain your primary goal as provided by the user. 
Reference portions of graph, coordinates, user hints, or anything else that may be useful.",
"relevant_graph": "List nodes or connections in the graph needed to complete your goal. 
If you need to explore, say unobserved_node(description). List ALL relevant nodes.",
"reasoning": "Explain how you are trying to accomplish this task in detail.",
"plan": "Your intended sequence of actions.", 
}
```
\end{minted}
\end{tcolorbox}
\end{listing*}

\begin{listing*}[h!]
\label{listing:perception}
\caption{LLM system prompt: perception API}
\begin{tcolorbox}[left=1mm, right=1.5mm, top=1.5mm, bottom=1mm] 
\footnotesize
\begin{minted}[]{python}
def remove(node: str) -> None:
    """Remove `node` and associated edges from graph."""


def add_node(type: str, name: str) -> None:
    """Add `node` of `type` to graph."""


def add_connection(type: str, node_1: str, node_2: str) -> None:
    """Add connection of `type` 
    (either `region_connection` or `object_connection`) between `node_1` and `node_2`."""


def update_robot_location(region_node: str) -> None:
    """Update robot's location in the graph to `region_node`."""


def update_node_attributes(region_node, **attributes) -> None:
    """Update node's attributes, where attributes are key-value pairs of attributes 
    and updated values."""


def no_updates() -> None:
    """There have been no updates."""
\end{minted}
\end{tcolorbox}
\end{listing*}

\begin{listing*}[h!]
\label{listing:planning_api}
\caption{LLM system prompt: planning API}
\begin{tcolorbox}[left=1mm, right=1.5mm, top=1.5mm, bottom=1mm] 
\footnotesize
\begin{minted}[]{python}
def goto(region_node: str) -> None:
    """Navigate to `region_node`."""


def map_region(region_node: str) -> List[str]:
    """Navigate to region in the graph and look for new objects.
    - region_node must be currently observed in graph and reachable from the robot's location.
    - This CANNOT be used to add connections in the graph.

    Will return updates to graph (if any).
    """


def extend_map(x_coordinate: int, y_coordinate: int) -> List[str]:
    """Try to add region node to graph at the coordinates (x_coordinate, y_coordinate).

    You should call this when your goal is far away (over 10 meters, for example).

    NOTE: if the proposed region is not physically feasible
    (because of an obstacle, for example), the closest feasible region will
    be added instead.

    Will return updates to graph (if any).
    """


def explore_region(region_node: str, exploration_radius_meters: float) -> List[str]:
    """Explore within `exploration_radius_meters` around `region_node`
    If (x, y) are the coordinates of `region_node` and `r` is the exploration radius.
    This will try to add regions at (x + r, y), (x - r, y), (x, y + r), (x, y - r).
    The robot will then map the discovered regions to find any unobserved objects.

    You should only call this if you are close to your goal (within exploration radius).

    Will return updates to graph (if any).
    """


def replan() -> None:
    """You will update your plan with newly acquired information.
    This is a placeholder command, and cannot be directly executed.
    """


def inspect(object_node: str, vlm_query: str) -> List[str]:
    """Gather more information about `object_node` by
    querying a vision-language model with `vlm_query`. Be concise in
    your query. The robot will also navigate to the
    region connected to `object_node`.

    Will return updates to graph (if any).
    """


def answer(answer: str) -> None:
    """Provide an answer to the instruction"""


def clarify(question: str) -> None:
    """Ask for clarification. Only ask if the instruction is too vague to make a plan."""
\end{minted}
\end{tcolorbox}
\end{listing*}

\begin{listing*}[h!]
\label{listing:advice}
\caption{LLM system prompt: planning advice}
\begin{tcolorbox}[left=1mm, right=1.5mm, top=1.5mm, bottom=1mm] 
\footnotesize
\begin{minted}[]{text}
The user given task with be prefaced by `task: `, and updates will be prefaced by `updates: `.

Remember the following when constructing a plan:
- You will receive feedback if your plan is infeasible. 
The feedback will discuss the problematic parts of your plan and reference specific regions of the 
graph. You will be expected to replan.

Remember the following at each planning iteration:
- When given an update, replan over the most recent instruction and updated scene graph.
- When given feedback, you must provide a plan that corrects the issues with your previous plan.


Planning Advice:
- Carefully explain your reasoning and all information used to 
create your plan in a step-by-step manner.
- Recall the scene may be incomplete. 
You may need to add regions or map existing regions to complete your task.
- Reason over connections, coordinates, and semantic relationships between objects and regions 
in the scene. For example, if asked to find a car, look near the roads.
- Coordinates are given west to east and south to north.

Before calling extend_map, consider this:
- If you need to find a path but there are NO existing connections, 
you should call extend_map in the direction of that region.
- Before you call extend_map ask: 
is there an existing connection I can use to get to my goal region? If so, use that.

Before calling explore_region, consider this:
- If you need to check if a path is clear, do not call explore. 
Rather, map the region to find obstacles.

Before calling goto, consider this:
- goto uses a graph-search algorithm to find an efficient path, 
so avoid calling goto on intermediate nodes.
- For example, if you path is ground_2 -> 
ground_7 -> ground_10. Call goto(ground_10) instead of goto(ground_7)
\end{minted}
\end{tcolorbox}
\end{listing*}

\begin{listing*}[h!]
\label{listing:in_context}
\caption{In context learning example provided to the LLM.}
\begin{tcolorbox}[left=1mm, right=1.5mm, top=1.5mm, bottom=1mm] 
\scriptsize
\begin{minted}[]{text}
EXAMPLE_GRAPH_1 = {
    "objects": [
        {"name": "house_1", "coords": [-1, -1]},
        {"name": "house_2", "coords": [-3, -1]},
        {"name": "grocery_store_1", "coords": [-5, -1]},
        {"name": "shed_1", "coords": [1, 3]},
        {"name": "shed_1", "coords": [1, 5]},
    ],
    "regions": [
        {"name": "example_road_1", "coords": [-1, 0]},
        {"name": "example_road_2", "coords": [-2, 0]},
        {"name": "field_11", "coords": [0, 1]},
        {"name": "field_13", "coords": [2, 3]},
    ],
    "object_connections": [
        ["house_1", "example_road_1"],
        ["house_2", "example_road_2"],
        ["shed_1", "field_11"],
        ["shed_2", "field_13"],
    ],
    "region_connections": [
        ["example_road_1", "example_road_2"],
        ["example_road_1", "field_11"],
        ["field_11", "field_13"],
    ],
    "robot_location": "example_road_1",
}

EXAMPLE_1 = [
    {
        "role": "user",
        "content": "task: I need a shovel. Is there one in the scene?"
        + f"Scene graph:{str(EXAMPLE_GRAPH_1)}",
    },
    {
        "role": "assistant",
        "content": '{\n\
        "primary_goal": "find a shovel for the user."\
        "relevant_graph": "field_11, field_13, unobserved_node(shovel)", \
        "reasoning": "The graph does not contain any shovels. 
        However, I know that the graph may be incomplete, so I will explore before providing a definitive answer. 
        I will first map, then if needed I will add regions. 
        There are two sheds in the scene, and shovels are often found near sheds. 
        Therefore, for each of the sheds, I will navigate to the nearby region and map.",\
        "plan": "[goto(field_11), map_region(field_11), goto(field_13), map_region(field_13)]"\n}',
    },
    {
        "role": "user",
        "content": "updates:[no_updates()]",
    },
    {
        "role": "assistant",
        "content": '{\n\
            "primary_goal": "find a shovel for the user."\
            "relevant_graph": "field_13, unobserved_node(shovel)",\
            "reasoning": "After mapping field_11, which is next to shed_1, I did not find a shovel. 
            I also planned on going to example_node_14 to map near shed_2, so I will continue my plan.",\
            "plan": "[goto(field_13), map_region(field_13)]"\n}',
    },
    {
        "role": "user",
        "content": "updates:[add_nodes({ name: shovel_1, type: object}), add_connections([shovel_1, field_13])]",
    },
    {
        "role": "assistant",
        "content": '{\n\
        "primary_goal": "find a shovel for the user."\
        "relevant_graph": "shovel_1", \
        "reasoning": "After exploring field_13, which is next to shed_2, I found a nearby shovel, shovel_1, 
        that is connected to field_13. This fulfills the users request.",\
        "plan": "[answer(There is a shovel, shovel_1, that is near shed_2 and connected to region_13.)]"\n}',
    },
    {
        "role": "user",
        "content": "task: How many grocery stores are currently in the scene?",
    },
    {
        "role": "assistant",
        "content": '{ \
        "primary_goal": "find a shovel for the user."\
        "relevant_graph": "grocery_store_1",\
        "reasoning": "There is one grocery store in the scene. Although the scene may be incomplete, 
        I am only asked about currently observed grocery stores. Thus the answer is one",\
        "plan": "[answer(There is one grocery store in the scene, grocery_store_1.)]"}',
    },
\end{minted}
\end{tcolorbox}
\end{listing*}

\clearpage

\section{Experimental details}
\label{sec:experiment_details}

This section provides details on the experimental tasks reported in Section~\ref{sec:experiments}.
We describe the mission, instruction given the SPINE, and the subtasks required. 
We then provide further discussion on experiments.

\subsection{Semantic Route inspection}

\noindent \textbf{Mission provided to SPINE}: ''There was a storm last night. I am worried that impacted logistics, because I need to drop off supplies today. Can I still do that?``

\shortskip

\noindent\textbf{Implied subtasks}: The planner must recognize that the delivery depot is the most likely place for supply delivery. The user wants to make sure the path between the current location and delivery depot is free. These subtasks are
\begin{enumerate}
    \item [1.] Recognize semantics. Primarily current location and delivery depot. Bonus: recognize that debris, puddles, fallen trees, etc, will give information about the extend of the storm. 
    \item [2.] Navigate along path path to delivery depot. At each step, if the robot cannot traverse an edge, it is likely blocked. 
\end{enumerate}

\begin{figure}[h!]
    \centering
    \includegraphics[width=0.99\linewidth]{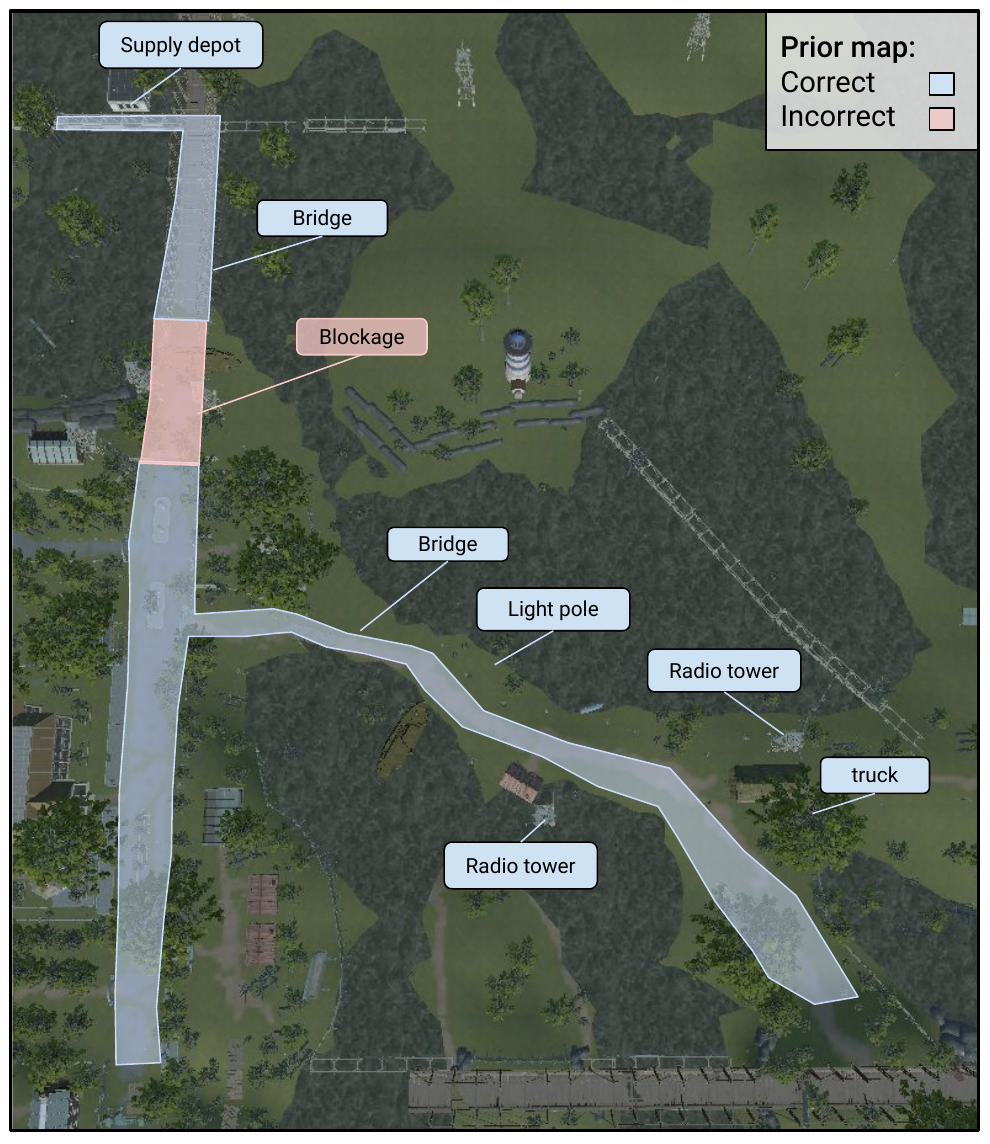}
    \caption{The semantic route inspection mission requires the planner to check if the path to the supply depot is free. Red is blocked by storm. The extent of prior is roughly 260m x 225m}
    \label{fig:sim-supply-1}
\end{figure}

\noindent\textbf{Map} is shown in Figure~\ref{fig:sim-supply-1}, which provided semantics:  ground, road, cabin, radio tower, truck, light pole, bridge, supply depot.

\begin{figure}[h!]
    \centering
    \includegraphics[width=0.99\linewidth]{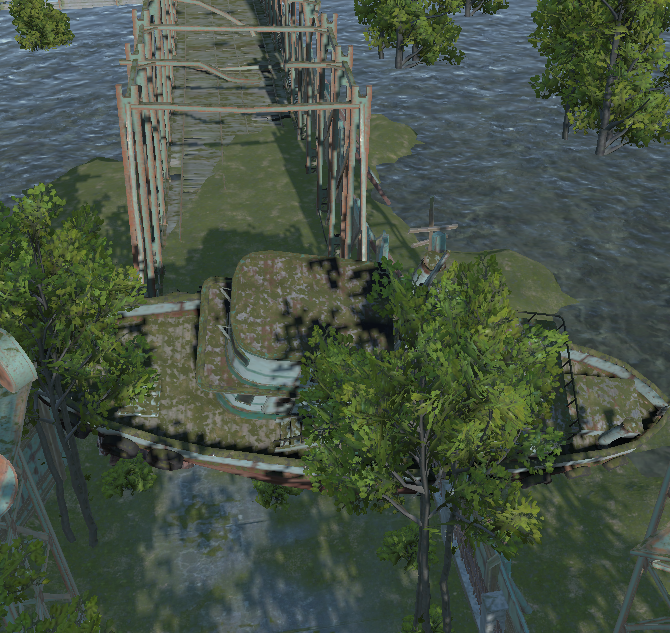}
    \caption{The semantic route inspection mission requires the robot to infer a path to the supply depot (across the bridge shown in the figure). During route inspection, the robot must recognize that the bridge is physically blocked.}
    \label{fig:sim-bridge-block-1}
\end{figure}

\subsection{Search and inspection with implicit goals}

\noindent\textbf{Mission provided to SPINE}: ''I sent a robot out to collect supplies from an incoming boat. I have not heard back. What happened?''

\shortskip

\noindent\textbf{Implied subtasks}: 
The planner must recognize that it is looking for a robot, and use the contextual information provided to infer the robot is likely near one of the three docks in the scene. The map does not provide a direct path to these docks, so the planner must explore in order to reach its goal locations. 
The planner must then find the mission robot, which is near the third dock. 
The implied subtasks are:
\begin{enumerate}
    \item [1.] Infer correct semantic labels (robot) and best search locations (three docks)
    \item [2.] Understand gaps in map (three major gaps)
    \item [3.] Navigate to the map boundary
    \item [4.] Extend map to the first dock
    \item [5.] Extend map to the second dock
    \item [6.] Extend map to the third dock
    \item [7.] Find and inspect robot
    \item [8.] Report findings to user
\end{enumerate}

\shortskip

\begin{figure}[h!]
    \centering
    \includegraphics[width=0.99\linewidth]{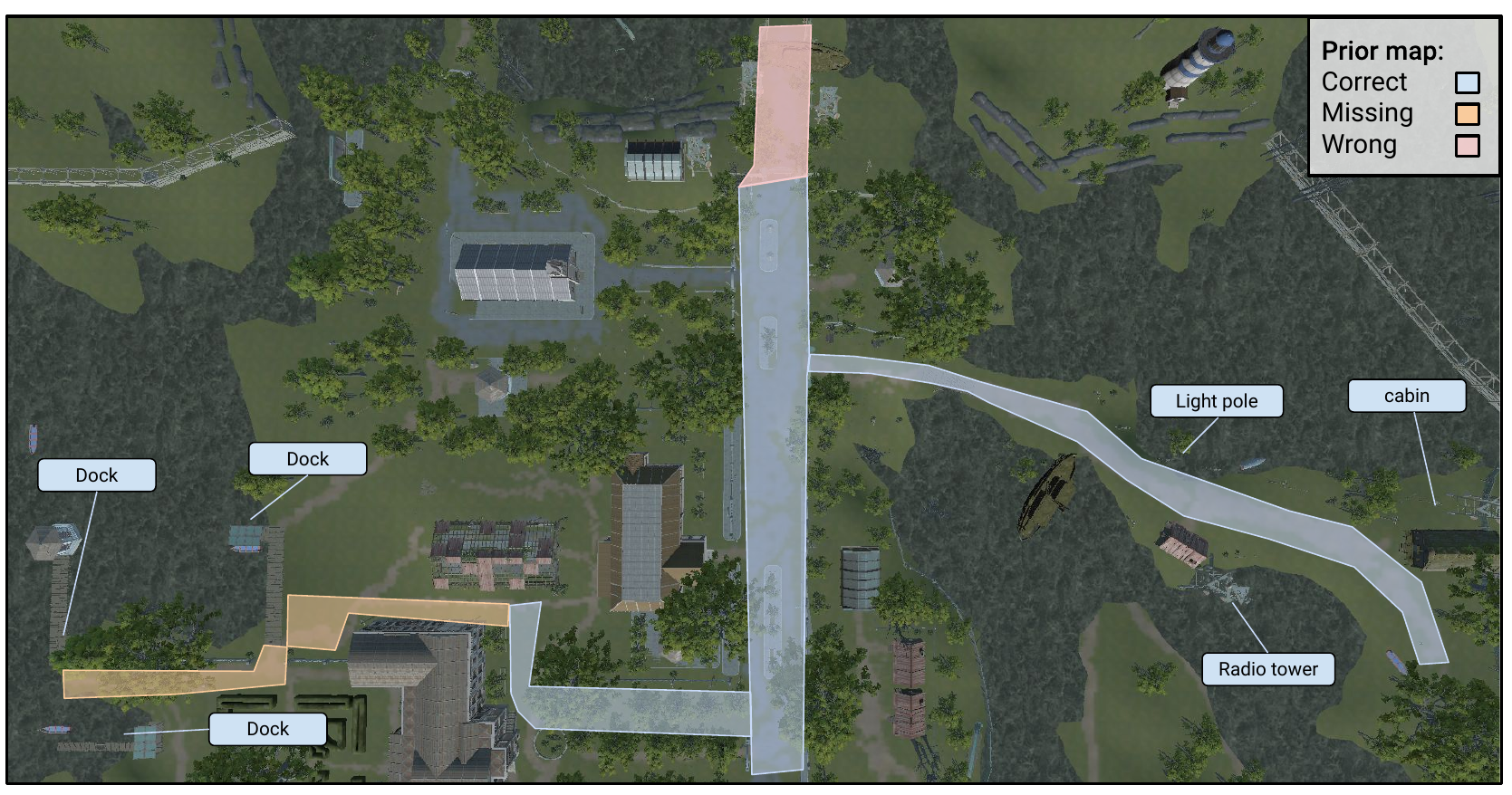}
    \caption{The search and inspection mission requires the planner to search near docks for another missing robot. The prior information provided to the planner and missing components in the map are illustrated.}
    \label{fig:supp-sim-2}
\end{figure}

\noindent\textbf{Map} is shown in Fig.~\ref{fig:supp-sim-2} with semantics dock, ground, road, cabin, radio tower, truck, light pole. 
Not all regions or semantics in prior are relevant to task.

\begin{figure}[]
    \centering
    \includegraphics[width=0.99\linewidth]{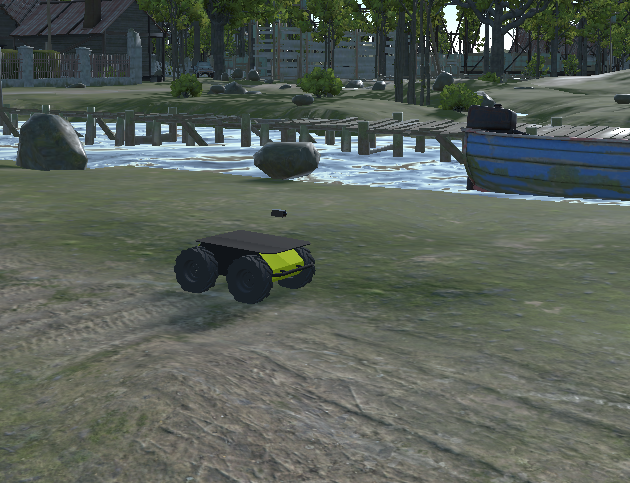}
    \caption{The search and inspection mission requires the planner to locate the robot shown in the figure and report the robot's position (eg ``robot is at location (x,y) and appears to be stationary''}
    \label{fig:enter-label}
\end{figure}

\subsection{Multi-object inspection with implied semantics}
\label{subsec:multi_object_search}

\noindent\textbf{Instruction provided to SPINE}: ''Communications are down. Can you figure out why?''

\shortskip

\noindent\textbf{Implied subtasks}: There are two radio towers provided in the prior map. The planner must infer that radio towers are relevant for communication, so it should inspect those. There is no direct path between the planners start locations and the radio towers, so the planner must explore.
The implied subtasks are:
\begin{enumerate}
    \item [1.] Identify inspection targets (radio towers)
    \item [2.] Go to region boundary
    \item [3.] Explore a path to the first radio tower
    \item [4.] Inspect the first radio tower by forming appropriate query (eg, ``is this radio tower damaged'') and reason over response
    \item [5.] Navigate to second radio tower 
    \item [6.] Inspect the first radio tower by forming appropriate query (eg, ``is this radio tower damaged'') and reason over response
    \item [7.] Provide information to user
\end{enumerate}

\shortskip

\begin{figure}[h!]
    \centering
    \label{fig:supp-sim-3}
    \includegraphics[width=0.99\linewidth]{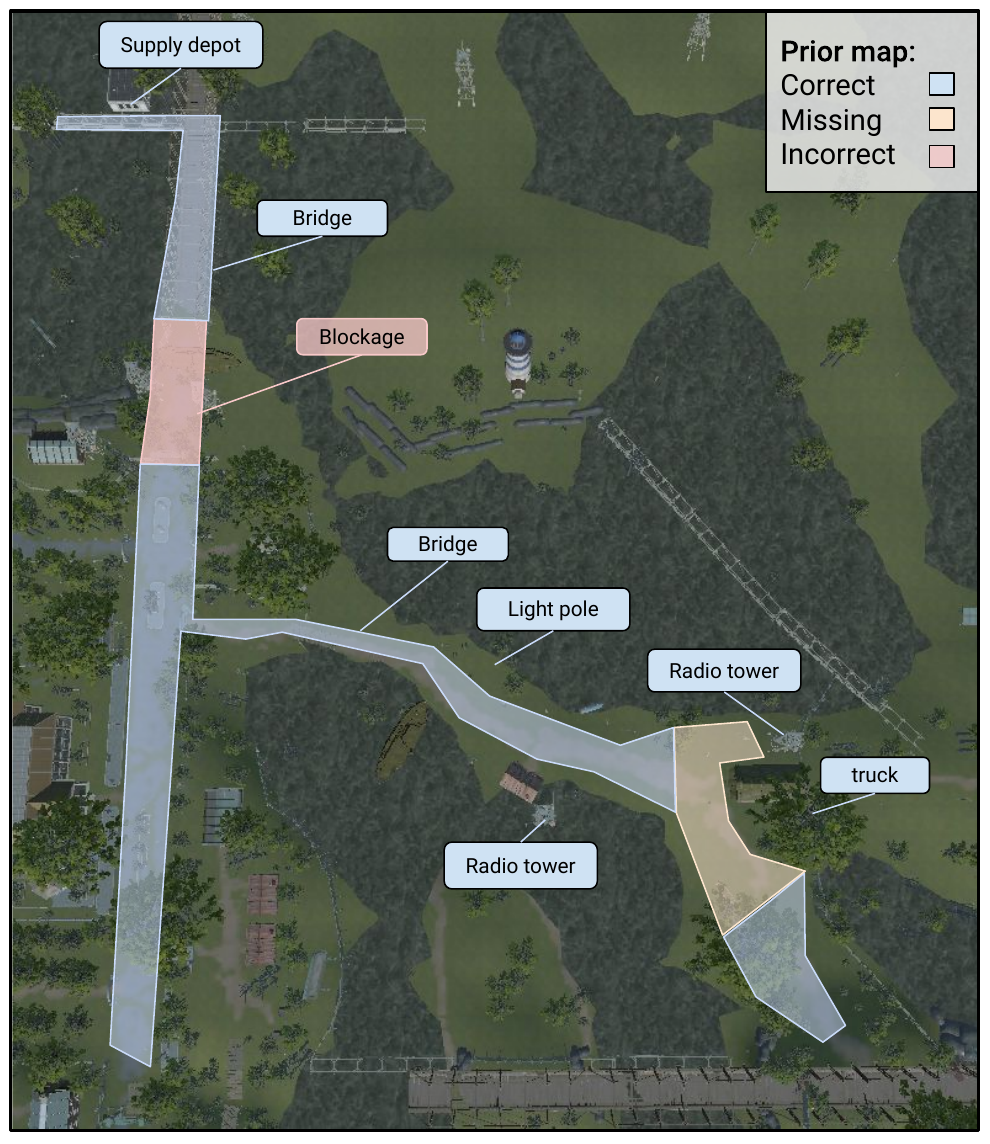}
    \caption{The multi-object inspection mission requires the planner to infer inspection targets (radio towers). There is no direct path provided in the prior, so the robot must explore to find a path. Furthermore, there are some distractors inclugin the bridge and supply depot (top of figure).}
    \label{fig:sim-3-map}
\end{figure}

\noindent\textbf{Map} is shown in Fig.~\ref{fig:supp-sim-3}, with provided semantics, ground, road, cabin, radio tower, truck, light pole.

\subsection{Semantic route inspection on real robot}

\noindent\textbf{Instruction given the SPINE} I am worried that recent construction on roads and fences impacted maritime supply logistics. Can you check? 

\shortskip

\begin{figure*}
    \centering
    \includegraphics[width=0.99\linewidth]{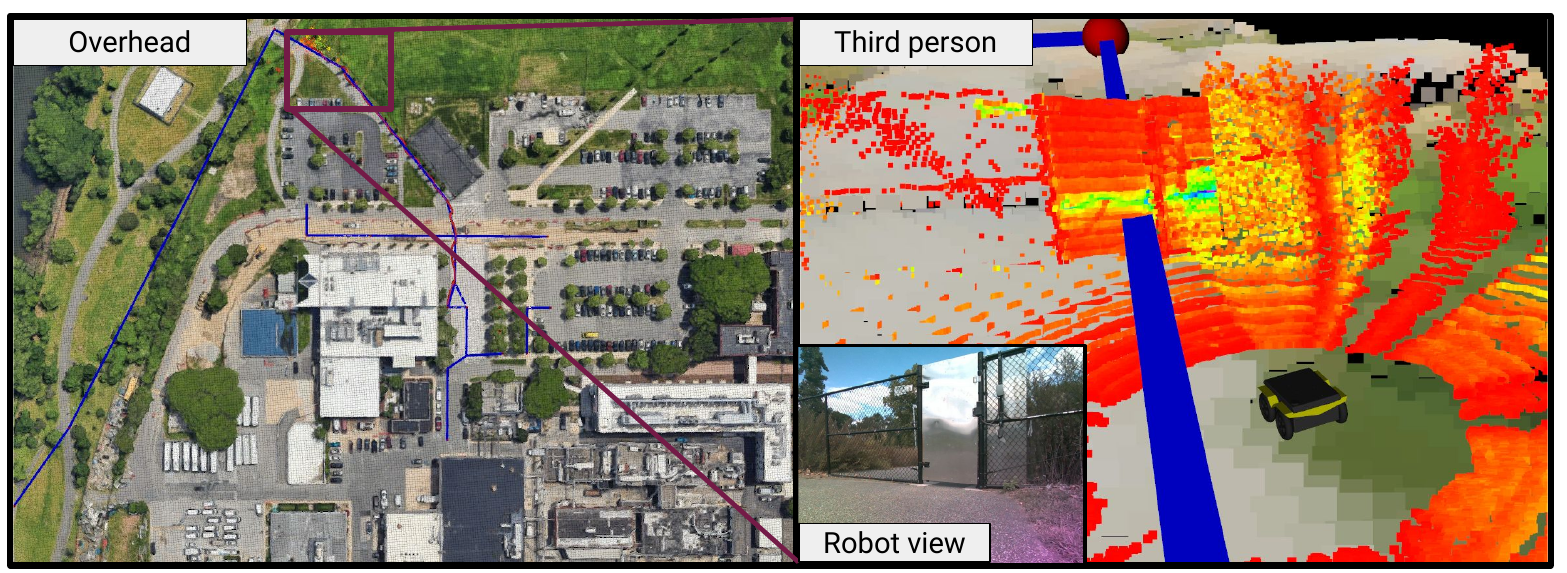}
    \caption{Example outcome on semantic route inspection mission. The mission implies that recent construction may have impacted the user's intended route to the dock (bottom left, off image). The planner searches along route until it finds a blockage (right). The planner then reports its findings to the user.}
    \label{fig:supp-dock-1}
\end{figure*}

\noindent\textbf{Implied subtasks:} The planner must recognize that the user is concerned about a path to the dock, which is provided in the prior. 
The prior is outdated; there is a newly built fence which obstructs the path. 
Furthermore, some of the path between the robots starting location and dock is missing. 
Thus, the planner must inspect the path towards the dock, recognized blockage, and report findings to the user.
A successful mission terminated when the the discovered the fence recently constructed, and the planner notifies the user.
See Fig.~\ref{fig:supp-dock-1}. 
The implied subtasks are:
\begin{enumerate}
    \item [1.] Specify correct semantics (roads, fences)
    \item [2.] Identify goal location (dock)
    \item [3.] Go to map boundary
    \item [4.] Fill in missing portion of path
    \item [5.] Use valid priors to navigate towards the dock
    \item [6.] Recognize blockage
    \item [7.] Report to user 
\end{enumerate}

\shortskip

\noindent\textbf{Map} is shown in Fig.~\ref{fig:exp_setup} with semantics courtyard, tree, parking lot, road, dock, path.

\subsection{Air-ground teaming on real robot}

\noindent\textbf{Mission provided to SPINE} You are assisting a high-altitude UAV in responding to an emergency chemical spill. Triage regions that are not visible from the air.

\shortskip

\noindent\textbf{Implied subtasks:} The planner must recognize that inside buildings and under trees cannot be observed from high-altitude UAVs, thus the planner should explore those regions.
There are regions of the map that are not provided in the prior, so the planner must explore.
The planner must also look for relevant semantics, including people and chemical barrels.
The implied subtasks are:
\begin{enumerate}
    \item [1.] Configure semantics (people, barrels)
    \item [2.] Go to the building entrance
    \item [3.] Explore to find a path inside 
    \item [4.] Recognize task-relevant objects
    \item [5.] Navigate to tree cover, which requires going to boundary of prior map
    \item [6.] Explore to tree cover
    \item [7.] Identify task-relevant objects.
\end{enumerate}

\shortskip

\noindent\textbf{Map} is shown in Fig.~\ref{fig:exp_setup} with semantics: parking lot, road, field, sidewalk, building, trees

\subsection{Discussion of results}
\label{subsec:further_disussion}

We observed comparative performance drop in SPINE (see Table~\ref{tab:agg-sim-results}) during multi-object inspection missions (Subsection~\ref{subsec:multi_object_search}). 
This mission required the planner to inspect two radio towers in the scene.
During some runs, the planner would inspect the first tower, learn that the tower was damaged, and terminate the mission. 
While this behavior is correct, it is not complete.

\begin{figure}[h!]
    \centering
    \includegraphics[width=0.99\linewidth]{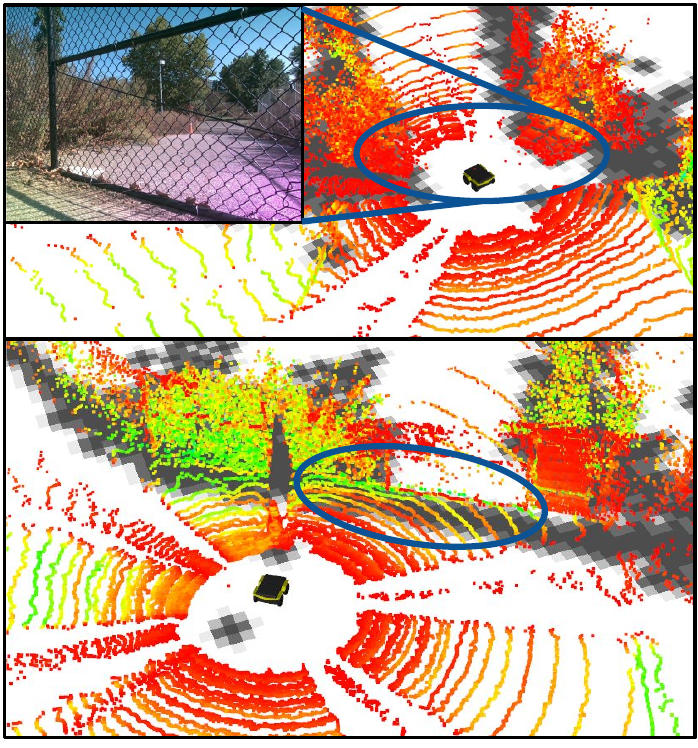}
    \caption{Cause of manual takeover during experiment. The LiDAR's minimum return distance was roughly 1 meter, so obstacles closer than this were not detected (top). When obstacles were not registered in the occupancy map, the robot tried to drive through them, which required manual takeover. The obstacles were picked back up again when robot moves farther away (bottom).}
    \label{fig:supp-takeover}
\end{figure}

For both the explicit tasking baseline and SPINE, there was one manual takeover for each experiment. 
These takeovers were both due to the minimum range of the obstacle detector, which was around 1 meters. 
If the robot came closer to one meter to an obstacle, that obstacle would not be registered in the perception costmap, thus the robot would try to drive into the obstacle. 
See Fig.~\ref{fig:supp-takeover} for an illustration.

\end{document}